\documentclass[10pt,twocolumn,letterpaper]{article}

\usepackage[accsupp]{axessibility}
\usepackage{iccv}              

\definecolor{iccvblue}{rgb}{0.21,0.49,0.74}
\usepackage[pagebackref,breaklinks,colorlinks,allcolors=iccvblue]{hyperref}

\usepackage{multirow}
\usepackage{multicol}
\usepackage{makecell}

\def\paperID{7641} 
\def\confName{ICCV}
\def\confYear{2025}

\title{Moto: Latent Motion Token as the Bridging Language for \\Learning Robot Manipulation from Videos} 
\author{\textbf{Yi Chen}$^{1*}$, \textbf{Yuying Ge}$^{2\dag}$, \textbf{Weiliang Tang}$^{3}$, \textbf{Yizhuo Li}$^{1,2}$, \textbf{Yixiao Ge}$^2$, \\\textbf{Mingyu Ding}$^4$, \textbf{Ying Shan}$^2$, \textbf{Xihui Liu}$^{1\dag}$\\
    $^1$The University of Hong Kong,
    $^2$ARC Lab, Tencent PCG,\\
    $^3$The Chinese University of Hong Kong,
    $^4$University of California, Berkeley\\
  \href{https://chenyi99.github.io/moto/}{\tt https://chenyi99.github.io/moto/}
 }

\begin{document}
\twocolumn[{
\renewcommand\twocolumn[1][]{#1}
\maketitle
\vspace{-20pt}
\begin{center}
    \captionsetup{type=figure}
    \includegraphics[width=1.0\textwidth]{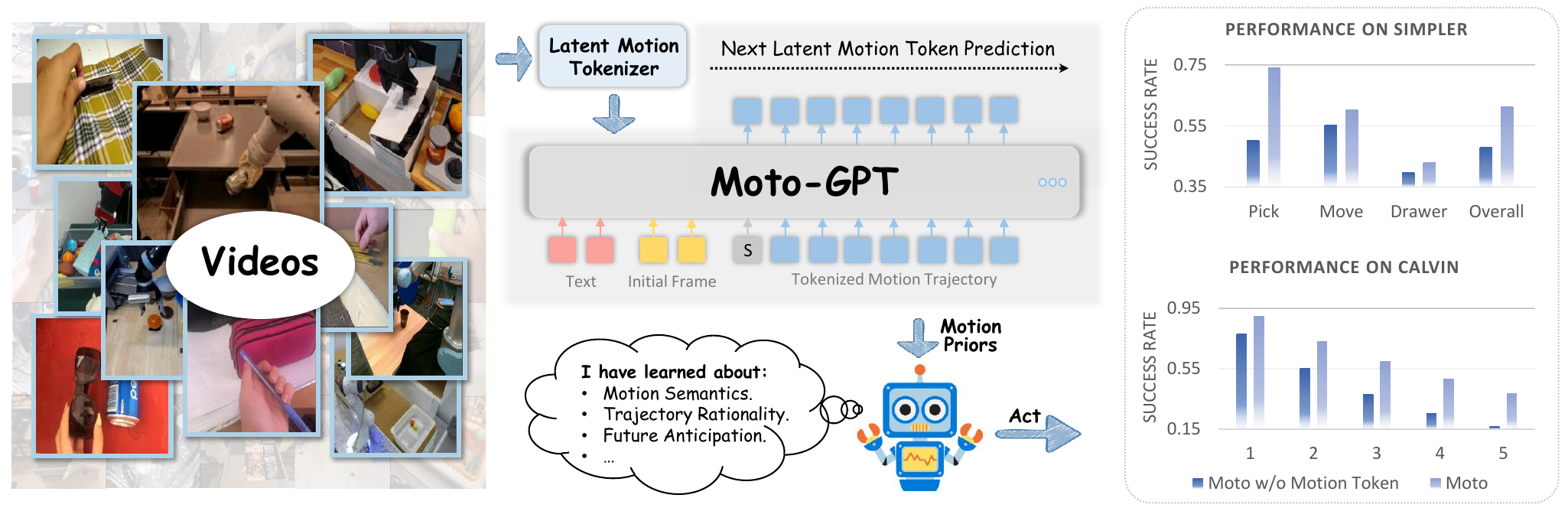}
    \vspace{-5pt}
    \captionof{figure}{The overview of \textbf{Moto}, which utilizes Latent \textbf{Mo}tion \textbf{To}kens as a bridging ``language'' for autoregressive pretraining on video data. The Moto-GPT pre-trained through next motion token prediction learns a wealth of motion-related prior knowledge from videos, which can be seamlessly transferred to enhance downstream robot manipulation tasks with significant performance gains.
    }
    \label{fig:teaser}
\end{center}
}]
\renewcommand{\thefootnote}{\fnsymbol{footnote}}
\setcounter{footnote}{1}
\footnotetext{Part of the work done during internship at Tencent ARC Lab.}
\setcounter{footnote}{2}
\footnotetext{Corresponding Authors.}
\renewcommand{\thefootnote}{\arabic{footnote}}
\setcounter{footnote}{0}
\begin{abstract}
Recent developments in Large Language Models (LLMs) pre-trained on extensive corpora have shown significant success in various natural language processing (NLP) tasks with minimal fine-tuning. 
This success offers new promise for robotics, which has long been constrained by the high cost of action-labeled data. We ask: given the abundant video data containing interaction-related knowledge available as a rich ``corpus'', \textbf{can a similar generative pre-training approach be effectively applied to enhance robot learning?} The key challenge is to identify an effective representation for autoregressive pre-training that benefits robot manipulation tasks.
Inspired by the way humans learn new skills through observing dynamic environments, we propose that effective robotic learning should emphasize motion-related knowledge, which is closely tied to low-level actions and is hardware-agnostic, facilitating the transfer of learned motions to actual robot actions. 
To this end, we introduce \textbf{Moto}, which converts video content into latent \textbf{Mo}tion \textbf{To}ken sequences by a Latent Motion Tokenizer, learning a bridging ``language'' of motion from videos in an unsupervised manner. 
We pre-train Moto-GPT through motion token autoregression, enabling it to capture diverse visual motion knowledge. After pre-training, Moto-GPT demonstrates the promising ability to produce semantically interpretable motion tokens, predict plausible motion trajectories, and assess trajectory rationality through output likelihood.
To transfer learned motion priors to real robot actions, we implement a co-fine-tuning strategy that seamlessly bridges latent motion token prediction and real robot control. Extensive experiments show that the fine-tuned Moto-GPT exhibits superior robustness and efficiency on robot manipulation benchmarks, underscoring its effectiveness in transferring knowledge from video data to downstream visual manipulation tasks.
\end{abstract}    
\section{Introduction}
\label{sec:intro}

Recent advancements in Natural Language Processing (NLP) have stemmed from successful autoregressive pre-training on large text corpora via next-word prediction~\citep{radford2018improving,brown2020language, ouyang2022training,touvron2023llama,dubey2024llama}. Pre-trained Large Language Models (LLMs) have shown exceptional performance across various downstream NLP tasks after fine-tuning on smaller datasets. This success opens new opportunity for robotics, which has been limited by the high costs of action-labeled data. Given the abundance of interaction-rich video data~\cite{bai2024sequential,yang2024video}, we ask: 
\textit{Can we leverage autoregressive pre-training on video data to improve robot learning?}

The main challenge is finding an appropriate representation for autoregressive pre-training on video data that effectively captures prior knowledge for robot manipulation. Pioneering research in video pre-training for robotics primarily focused on static frames, emphasizing frame-level visual details~\cite{wu2024unleashing,cheang2024gr,escontrela2023video}. However, humans learn skills by observing dynamic environments, focusing on changes in state—what we term motion. Thus, we argue that effective autoregression for robotics should \textit{prioritize motion-related knowledge}, which aligns closely with low-level robot actions and is hardware-agnostic, facilitating the transfer of learned motions to actual robot actions through fine-tuning.

In this work, we introduce \textbf{Moto}, which utilizes Latent \textbf{Mo}tion \textbf{To}kens as a bridging ``language'' to model visual motions between video frames in an unsupervised manner. As illustrated in Fig.~\ref{fig:teaser}, we first train a discrete Latent Motion Tokenizer to produce compact latent motion tokens that capture dynamics between video frames without external supervision. We then pre-train Moto-GPT using a GPT-based architecture to predict the next latent motion token, absorbing motion priors from videos. These learned priors are subsequently transferred to enhance robot manipulation performance through a co-fine-tuning strategy.

\begin{figure*}[t]
    \centering
    \includegraphics[width=1.0\textwidth]{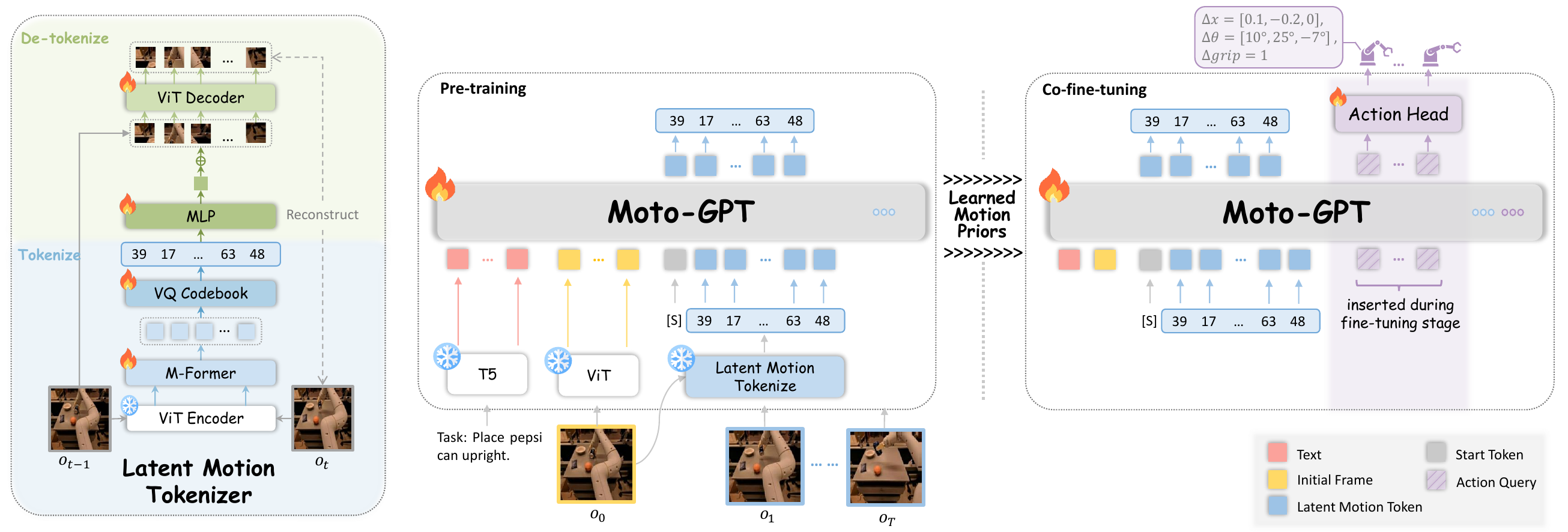}
    \vspace{-20pt}
    \caption{
    Overview of Moto's three training stages: (1) The Latent Motion Tokenizer encodes key visual motions between video frames into compact latent tokens in an unsupervised manner using pure video data. (2) Moto-GPT is pre-trained with autoregressive motion token prediction to learn motion priors from video-instruction pairs. (3) Moto-GPT is co-fine-tuned on action-labeled trajectories to predict robot actions based on the output of learnable action query tokens while maintaining the next-motion-token prediction objective.}
    \label{fig:model_overview}
    \vspace{-10pt}
\end{figure*}

Specifically, as shown in Fig.~\ref{fig:model_overview}, the Latent Motion Tokenizer encoder employs a VQ-VAE-based architecture~\cite{van2017neural} to compress two successive video frames into discrete tokens. 
By regularizing the decoder to reconstruct the second frame from the first frame and the tokens, the tokenizer is trained to effectively capture the changes between video frames, which often arise from motion.
Once the tokenizer is trained, we obtain latent motion tokens of every two consecutive frames in a video clip and concatenate them into a sequence to represent the motion trajectory.
Subsequently, Moto-GPT is pre-trained on these sequences by predicting the next token based on the initial frame and corresponding language instruction. After this pre-training phase, Moto-GPT is capable of generating plausible trajectories by predicting latent motion tokens autoregressively. 

To adapt Moto-GPT for downstream robot manipulation tasks, we concatenate action query tokens with latent motion token chunk at each time step for co-fine-tuning on action-labeled robot data. The action query tokens are processed by a learnable module to predict low-level actions, while the motion tokens are fine-tuned using the original next-token prediction mechanism. This co-fine-tuning strategy effectively transfers abstract intentions in learned motion priors into precise action execution, allowing the model to utilize the inherent knowledge of the pre-trained Moto-GPT for successful manipulation.

We conduct extensive experiments to validate our claims from various perspectives: (1) \textbf{Latent Motion Tokens as an Interpretable Motion Language}: 
Experiments show that latent motion tokens encapsulate compact and expressive representations of visual motions and exhibit promising cross-embodiment transfer ability (even from human to robot). 
(2) \textbf{Pre-trained Moto-GPT as a Useful Motion Prior Learner}: Results indicate that pre-trained Moto-GPT achieves promising outcomes in predicting plausible motion trajectories and assessing the rationality of robot trajectories based on output likelihood. (3) \textbf{Fine-tuned Moto-GPT as an Effective Robot Policy}: The fine-tuned Moto-GPT demonstrates significant performance gains over counterparts trained from scratch without motion priors, especially with limited training data.
The performance can be further boosted with human video pre-training, highlighting the potential of our approach in transferring motion knowledge learned from Internet-scale videos to robot manipulation.

In summary, our contributions are 
as below:
\begin{itemize}
    \item Introduction of Latent Motion Tokens, which model visual motions between video frames in an unsupervised manner, serving as a bridging ``language'' for autoregressive pre-training to enhance robot learning.
    \item Pre-training of Moto-GPT through next latent motion token prediction on video data, enabling the model to learn useful motion priors without requiring action annotations.
    \item Implementation of a co-fine-tuning strategy to successfully transfer learned motion priors to actual robot manipulations, with the fine-tuned model showing competitive performance on robotic benchmarks.
    \item We conduct systematic experiments and analyses to validate each training stage of our approach and the effectiveness of the resulting robot policy. Our open-source code provides clear guidance for future exploration.
\end{itemize}

\section{Related Work}

\noindent \textbf{Vision-Language-Action Models.}
Recent studies have utilized transformers as vision-language-action (VLA) architectures to generate robot actions from observations and language instructions~\cite{brohan2022rt,reed2022a,jiang2023vima,li2024cogact}. VLA model pre-training has gained traction, with approaches either fine-tuning policy models from powerful vision-language models pre-trained on large image-text datasets~\cite{driess2023palm,zitkovich2023rt,li2024visionlanguage} or training generalist models on diverse robot data with action labels~\cite{mees2024octo,kim2024openvla,doshi2024scaling,vuong2023open,black2024pi_0}. Our work enhances VLA models through generative pre-training on video data, providing richer interaction details.
Beyond VLA models, several studies improve robot manipulation by incorporating multi-perspective views and depth information~\cite{liu2024robouniview,bu2024closedloop,zhen20243d}, and employing techniques like action chunking and policy diffusion for precision~\cite{haldar2024baku,chi2023diffusion,ke2024d}. Additionally, some works~\cite{garg2022lisa,liang2024skilldiffuser} decompose high-level language instructions into latent skills via auxiliary training objectives.

\noindent\textbf{Robot Learning from Videos.} 
Early works~\cite{nair2023r3m,ma2023vip} used contrastive learning with egocentric videos to enhance visual representations. Some studies~\cite{du2024learning,ko2024learning,liang2024dreamitate,black2024zeroshot,li2024gr} generate videos as intermediate plans for low-level control. Recent research~\cite{wu2024unleashing,cheang2024gr,he2024learning} has focused on generative video pre-training for end-to-end policy models. \citet{escontrela2023video} pre-train autoregressive video prediction models for reinforcement learning.  These works primarily use pixel values or patch-level tokens. In contrast, our approach targets latent motion tokens, emphasizing key visual motions.
Some studies build world models through action-conditioned video generation~\cite{Hafner2020Dream,yang2024learning,wu2024ivideogpt}, aiding reinforcement learning.  Genie~\cite{bruce2024genie} introduces unsupervised learning of latent actions from large-scale videos for 2D gaming simulators. Similar latent action concepts are explored in LAPO~\cite{schmidt2023learning} and DynaMo~\cite{cui2025dynamo}, though LAPO is limited to video games and DynaMo focuses on visual representation learning.
Concurrently, LAPA~\cite{ye2024latent} is pre-trained to predict one-step future latent action, while IGOR~\cite{chen2024igor} uses latent actions as intermediate goals. 
Our work differs by pre-training an end-to-end policy model to autoregressively predict a trajectory of latent motion tokens for future video clips, providing a more natural modeling of sequential visual motions from videos. Additionally, we conduct a rigorous analyses to validate the effectiveness of policy and each training stage, offering clear guidance for future research.

\section{Methodology}
\subsection{Overview}

Moto utilizes autoregressive generative pre-training on latent motion token sequences to learn motion priors from videos, followed by co-fine-tuning on action-labeled data for robot control.
As illustrated in Fig.~\ref{fig:model_overview}, Moto consists of three stages: 1) unsupervised training of the Latent Motion Tokenizer, 2) pre-training of the generative model Moto-GPT, and 3) co-fine-tuning for robot action policy.
In Sec.~\ref{sec:motion_tokenizer}, we detail the Latent Motion Tokenizer, which encodes visual dynamics into quantized latent motion tokens. We also describe the training procedures for Moto-GPT, including motion token autoregressive pre-training in Sec.~\ref{sec:pre-training} and supervised co-fine-tuning in Sec.~\ref{sec:fine-tuning}. Implementation details can be found in the Supplementary Material.

\subsection{Latent Motion Tokenizer}\label{sec:motion_tokenizer}
The Latent Motion Tokenizer, as shown in Fig.~\ref{fig:model_overview}, learns a latent ``language'' to capture essential visual motions between successive video frames\footnote{To ensure significant visual differences, we down-sample the original video by a certain rate.} in an unsupervised manner. 
The architecture follows a standard auto-encoder design for motion tokenization and detokenization.
The tokenization employs an M-Former, a multi-layer transformer that extracts motion features from the last-layer patch features of the current frame \(o_t\) and the preceding frame \(o_{t-1}\) using a frozen pre-trained ViT encoder~\cite{he2022masked}. We concatenate 8 learnable query embeddings with these patch features as additional input to the M-Former, where the queries interact through self-attention layers. The output query features are then processed by a VQ codebook with a vocabulary size of 128 to produce discrete latent motion tokens.

For de-tokenization, we use a ViT Decoder for image reconstruction, which takes the linearly embedded patches of \(o_{t-1}\) and recovers the pixel values for \(o_t\) based on the latent motion tokens. An MLP projects the concatenated quantized embeddings of the latent motion tokens into a compact embedding (1 token), which is added to each input patch embedding. 
This conditional embedding acts as an information bottleneck between the encoder and decoder, enabling the ViT Decoder to capture nuanced changes between frames and accurately transform \(o_{t-1}\) into \(o_t\).

The components of the Latent Motion Tokenizer are jointly optimized using the standard VQ-VAE objective~\cite{van2017neural}, which includes reconstruction loss, vector quantization loss, and commitment loss. We specifically use the MSE loss between the output pixel values from the ViT Decoder and the ground-truth pixel values of \(o_t\) as the reconstruction loss. Once trained, the Latent Motion Tokenizer is frozen to produce unified sequential motion representations for videos through ``bi-frame'' tokenization. Additionally, with the initial observation and specified latent motion tokens, the decoder can function as a ``simulator'' to generate rollouts for visualizing environmental changes.

\subsection{Motion Token Autoregressive Pre-training}\label{sec:pre-training}
With the Latent Motion Tokenizer, Moto-GPT is allowed to learn about diverse visual motions from videos, using latent motion tokens as a bridging language.
As shown in Fig.~\ref{fig:model_overview}, Moto-GPT is pre-trained with a next-motion-token prediction objective. For a video clip $[o_0, o_1, ..., o_T]$, we derive a chunk of latent motion tokens for each pair of consecutive frames, concatenating them chronologically to form a sequence. Moto-GPT employs a GPT-style transformer for autoregression on these motion token trajectories. Additionally, we prepend the text features from the instruction and the visual features from the initial video frame as input prompts. The pre-training objective maximizes the likelihood of the ground-truth latent motion token sequence given the language instruction and the initial video frame:
\begin{equation}
    \mathcal{L}_{motion} = -\sum_{i=1}^{M} \log P(m_i | \boldsymbol{l}, \boldsymbol{v},  \boldsymbol{m}_{<i}; \boldsymbol{\Theta}),
\end{equation}\label{eq:autoregressive_likelihood}  
where $\boldsymbol{l}$ and $\boldsymbol{v}$ are text and visual features from the frozen pre-trained T5~\cite{raffel2020exploring} and ViT~\cite{he2022masked} models, respectively. $\boldsymbol{m}_{<i}$ represents the latent motion tokens preceding the current token $m_i$, and $\boldsymbol{\Theta}$ denotes the trainable model parameters. Here, $M=K*T$, where $K$ is the number of tokens for motion between successive frames and $T$ is the video length.

\subsection{Co-fine-tuning for Robot Manipulation}\label{sec:fine-tuning}

After pre-training, Moto-GPT can anticipate future trajectories by generating latent motion tokens based on language instructions and initial observations. 
This process resembles the policy inference of real robots if we take the codebook of latent motion tokens as an abstract action space. 
However, a gap remains in achieving precise robot control.

To address this, 
during fine-tuning, we introduce special action query tokens into Moto-GPT's input, enabling the generation of real robot actions through a flexible action head, as illustrated in the right part of Fig.~\ref{fig:model_overview}. Specifically, $N$ query tokens are added after the latent motion token chunk at each time step, where $N$ corresponds to the number of robot actions occurring between two video frames. 
The fine-tuning stage follows the same causal mask mechanism as pre-training in general.
Nevertheless, the latent motion tokens do not attend to the newly inserted action query tokens to stay consistent with the pre-training setting.
Besides, we randomly mask 50\% of the attention from action query tokens to latent motion tokens, allowing knowledge transfer while reducing dependency on ground-truth conditions.
This also improves inference efficiency, enabling direct queries to Moto for real actions without generating latent motion tokens. This can be achieved by using padding tokens as placeholders for latent action tokens, blocking attention from action query tokens to these placeholders.

\begin{figure}[!t]
    \centering
    \includegraphics[width=0.45\textwidth]{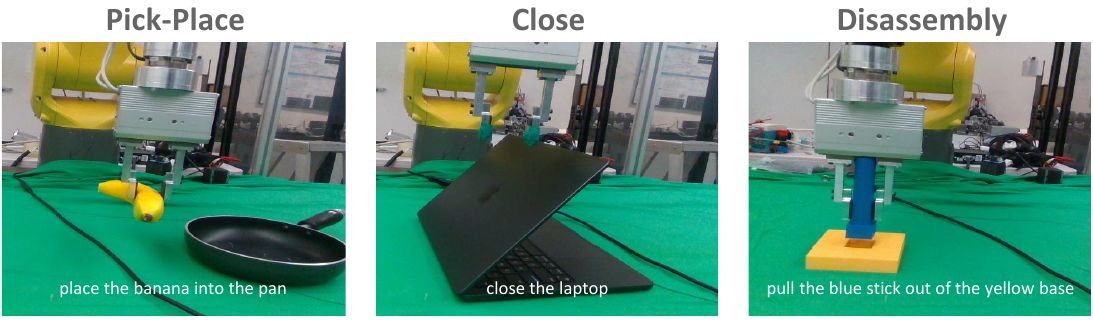}
    \vspace{-5pt}
    \caption{
    Illustration of real-world evaluation tasks.
    }
    \vspace{-10pt}\label{fig:real_robot_illustration}
\end{figure}

An MLP-based action head projects the output hidden state of each action query token into the real robot action space. We apply Smooth-L1 loss for continuous action components, such as positional ($\Delta x$) and rotational ($\Delta \theta$) displacements, and Binary Cross Entropy (BCE) loss for binary components, like the gripper's open/close state ($\Delta grip$)\footnote{The action space may vary with different robot embpdiments. For example, the Google Everyday Robot uses a continuous value for gripper extension, necessitating Smooth-L1 loss for $\Delta grip$.}. The total action loss $\mathcal{L}_{action}$ is defined as:
\begin{equation}
    \mathcal{L}_{action} = \mathcal{L} (\Delta x) + \mathcal{L} (\Delta \theta) + \mathcal{L} (\Delta grip)
\end{equation}
We retain the training objective for latent motion token prediction to ensure Moto-GPT retains the motion priors learned from videos. Thus, the overall loss function for the fine-tuning stage is:
\begin{equation}
    \mathcal{L}_{ft} = \mathcal{L}_{motion} + \mathcal{L}_{action}
\end{equation}

\begin{figure*}[!t]
    \centering
    \includegraphics[width=1.0\textwidth]{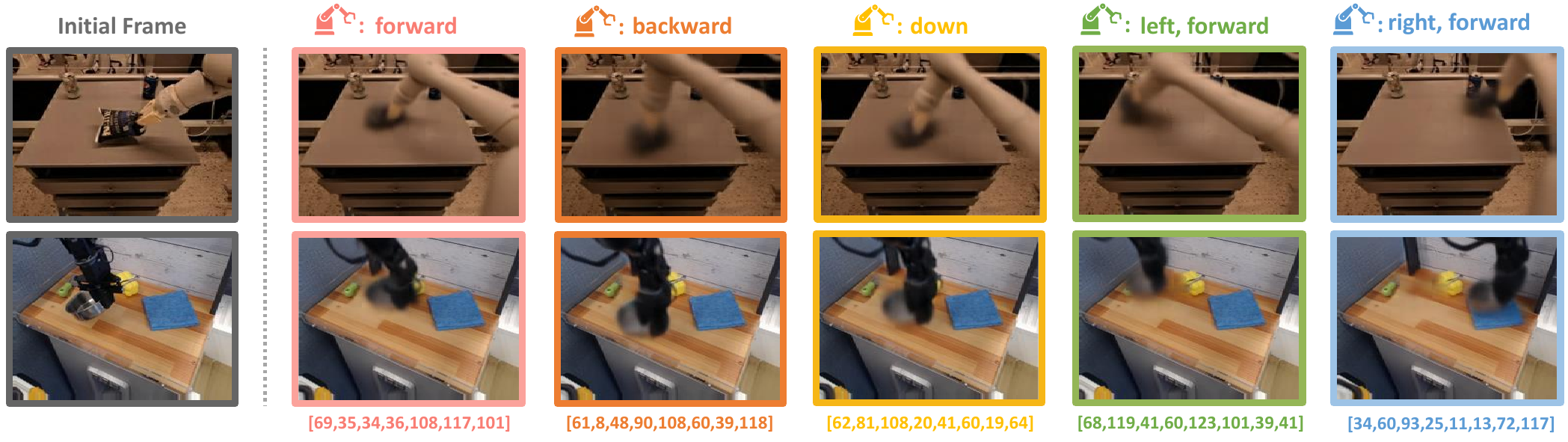}
    \vspace{-20pt}
    \caption{
    Interpretability of latent motion tokens. Each row displays reconstructed frames from the same initial frame using different latent motion tokens, while each column shows frames reconstructed from the same latent motion tokens with varying initial frames. The latent motion tokens exhibit consistent (see columns) and discriminative (see rows) semantics, despite being trained in an unsupervised manner.
    }
    \label{fig:controllability}
\end{figure*}

\begin{figure*}[!t]
    \centering
    \includegraphics[width=1.0\textwidth]{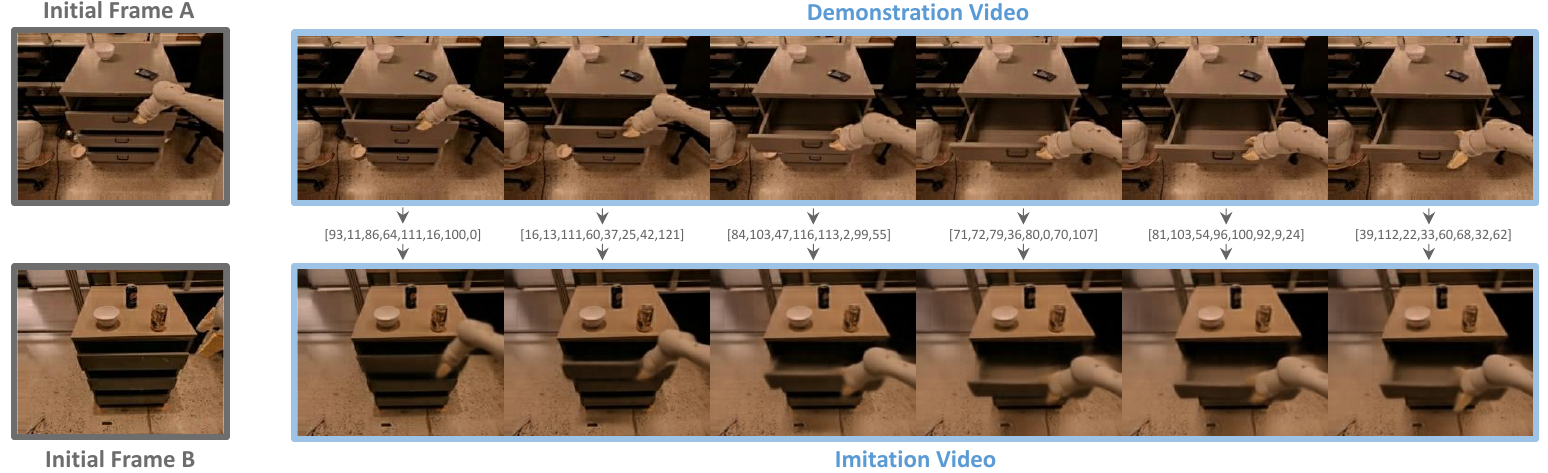}
    \vspace{-17pt}
    \caption{
    Video imitation generation via latent motion tokens, 
     where a sequence of motion tokens extracted from a demonstration video are decoded into a new video. This generated video is based on a different initial frame while preserving the original movement semantics.
    }
    \vspace{-7pt}
    \label{fig:demo_transfer}
\end{figure*}

\section{Benchmarks and Datasets}

More details of experimental settings can be found in the supplementary material.

\noindent \textbf{SIMPLER~\cite{li2024evaluating}.} We focus on three tasks with the Google Everyday Robot: ``pick coke can'', ``move near'', and ``open/close drawer''. For pre-training, we leverage a subset of Open-X-Embodiment (OXE)~\cite{vuong2023open}, consisting of 109k trajectory videos~\cite{brohan2022rt,walke2023bridgedata,rosete2022tacorl,mees23hulc2,dass2023jacoplay,luo2023multistage,mandlekar2019scaling,pari2021surprising,zhu2022viola,BerkeleyUR5Website,zhou2023train}. Fine-tuning is performed using 73k action-labeled expert trajectories from the RT-1 Robot-Action dataset~\cite{brohan2022rt}.

\noindent \textbf{CALVIN (ABC$\longrightarrow$D)~\cite{mees2022calvin}.} We assess long-horizon task completion with the Franka Emika Panda robot, requiring consecutive completion of 5 out of 34 tasks in each trial in a zero-shot setting. Pre-training involves all play videos from environments A, B, and C, with 35\% (18k videos) containing language annotations. Fine-tuning uses 18k expert trajectories with action labels from the same environments. Testing is conducted in the unseen environment D.

\noindent \textbf{Real-world Robot Experiments.} We conduct real-world evaluations with a FANUC LR Mate 200iD robot on three tasks: ``pick-place banana'', ``close laptop'', and ``disassembly'' (Fig.~\ref{fig:real_robot_illustration}). Pre-training utilizes OXE data, and we collect 90 teleoperated demonstrations (30 per task) for fine-tuning. Each task is tested 10 times with randomized object positions. Generalizability is evaluated in two scenarios: (i) Novel Object, altering the object's color, texture, and shape; (ii) Visual Distractor, introducing irrelevant objects.

\section{Experiments}
To comprehensively evaluate the effectiveness of Moto, we study three key experimental questions:
\begin{itemize}
\item \textbf{Q1 (Interpretability):} Does the Latent Motion Tokenizer learn interpretable latent motion tokens that effectively represent visual motions from videos?
\item \textbf{Q2 (Motion Priors):} Does Moto-GPT gain meaningful prior knowledge of motion trajectories through autoregressive pre-training on latent motion token sequences?
\item \textbf{Q3 (Performance):} Can the motion priors be transferred to enhance policy performance in robot manipulation benchmarks through efficient fine-tuning?
\end{itemize}

\subsection{Latent Motion Token as an Interpretable Motion Language}

The Latent Motion Tokenizer effectively reconstructs authentic next frames using ground-truth latent motion tokens, capturing key dynamics between initial and subsequent frames, as evidenced by supplementary examples. This demonstrates its ability to represent fine-grained motion details, with its decoder acting as a reliable simulator for visualizing environmental changes. 
Fig.~\ref{fig:controllability} highlights the controllability and consistency of latent motion tokens. Different token chunks produce varied motion orientations and scales, while identical chunks yield consistent effects across different starting observations with varied robot embodiments. By concatenating token chunks from consecutive frames, motion trajectories can be represented sequentially, akin to natural language, enabling contextualized motion generation across diverse initial observations (Fig.~\ref{fig:demo_transfer}). This underscores the potential of latent motion tokens as a unified language for imitation learning.

Quantitatively, Table~\ref{tab:semantic_acc} shows the semantic interpretability of latent motion tokens. A video classifier using initial frame ViT patch features and concatenated latent motion tokens for seven subsequent frames achieves 79.7\% accuracy in predicting semantic labels for 34 CALVIN tasks. This performance is comparable to using ViT features for all eight frames, despite reducing input features from 196 to 8 tokens per frame, confirming that latent motion tokens provide a compact, expressive, and interpretable representation of visual motions linked to high-level semantics.

\subsection{Moto-GPT as a Useful Motion Prior Learner}
\begin{figure}[!t]
    \centering
    \includegraphics[width=0.45\textwidth]{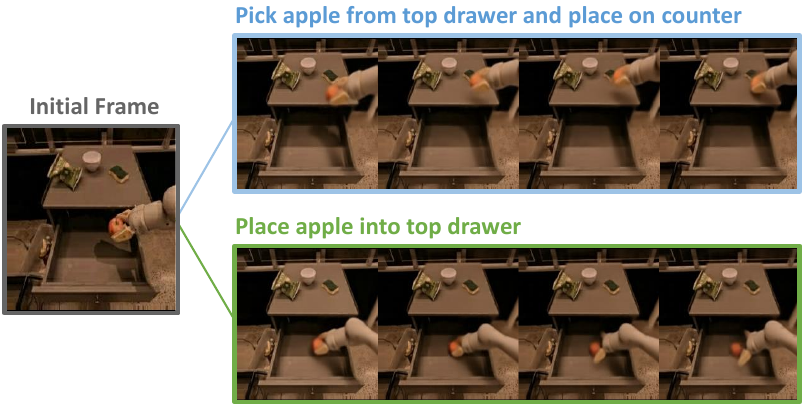}
    \caption{
    Visualization of video trajectories generated from a sequence of latent motion tokens, which are predicted by the pre-trained Moto-GPT given different language instructions.
    }
    \vspace{-5pt}
    \label{fig:motion_prediction}
\end{figure}

\begin{figure}[!t]
    \centering
    \includegraphics[width=0.4\textwidth]{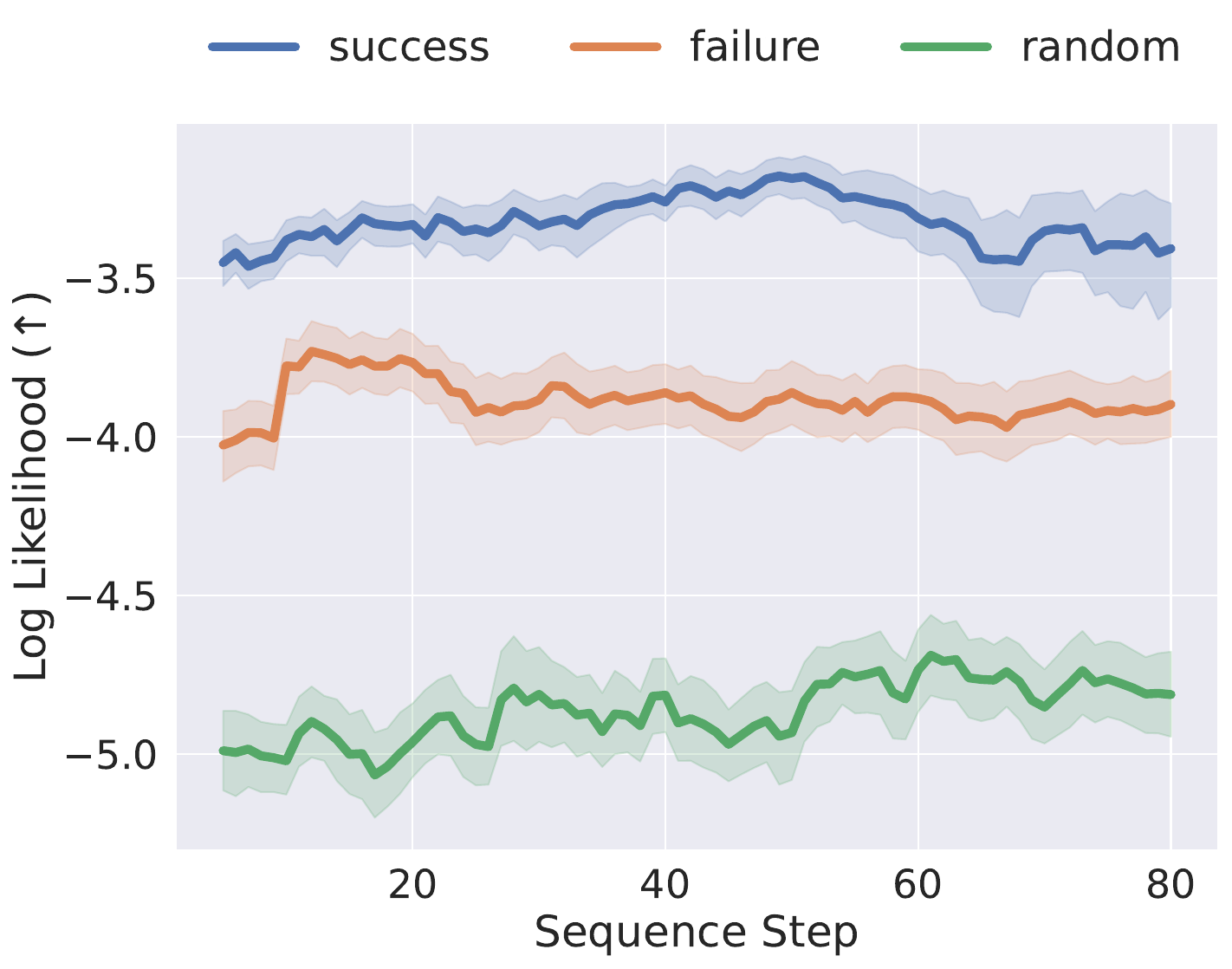}
    \vspace{-5pt}
    \caption{
    Moto-GPT distinguishes successful, failed, and random robot trajectories using log-likelihoods, enabling effective assessment of trajectory rationality and potential reward signals.
    }
    \label{fig:trajectory_rationality}
    \vspace{-10pt}
\end{figure}

\begin{table}
\caption{
Video classification accuracy with varied representations.
}
\small
    \centering
    \begin{tabular}{l c}
    \toprule
    Video Representation & Semantic Acc. \\
    \midrule
    Initial frame & 0.292 \\
    Initial frame repeated by 8 times & 0.283 \\
    Initial frame + 7 subsequent frames & 0.828\\
    \makecell{Initial frame + 7 latent motion token chunks} & 0.797 \\
    \bottomrule
    \end{tabular}\label{tab:semantic_acc}
    \vspace{-7pt}
\end{table}

Moto-GPT is pre-trained through autoregression on videos using latent motion tokens, enabling it to predict motion trajectories from initial observations and diverse language prompts, as shown in Fig.~\ref{fig:motion_prediction}.  The top-k token prediction accuracy and the visualization of predicted video trajectories for more complex actions (e.g., ``rotation'' and ``stacking'') are provided in the supplementary material. 
This demonstrates Moto-GPT's ability to acquire prior knowledge essential for robot action inference based on human instructions.
Additionally, latent motion tokens allow Moto-GPT to interpret trajectory videos as compact token sequences and evaluate their rationality through the autoregressive likelihood (Eq.~\ref{eq:autoregressive_likelihood}). Fig.~\ref{fig:trajectory_rationality} illustrates the potential of using Moto's log-likelihoods as a reward signal for trajectory videos, indicating how well a trajectory aligns with Moto-GPT's distribution and measuring the temporal consistency of behavior.
To assess this, we collected 98 video triplets in CALVIN using the baseline policies and a random policy. Each triplet consists of three types of trajectory videos originating from the same environment state. The averaged log-likelihoods for each trajectory type at each sequence step, shown in Fig.~\ref{fig:trajectory_rationality}, clearly differentiate successful trajectories from failures and random attempts.

\subsection{Moto-GPT as an Effective Robot Policy} 

\begin{table*}
\caption{SIMPLER evaluation results of models pre-trained on Open-X-Embodiment~\cite{vuong2023open} datasets. The ``Overall'' column reports the success rate averaged across the sub-tasks of all task types.}
\small
\centering
\begin{tabular}{l c c c c c c c c c}
\toprule
\multirow{2}{*}{Method} & \multicolumn{4}{c}{Pick Coke Can} & Move Near & \multicolumn{3}{c}{Open / Close Drawer} & Overall \\
\cmidrule(r){2-5}  \cmidrule(rl){6-6} \cmidrule(rl){7-9} \cmidrule{10-10}
& \makecell{Horizontal} & \makecell{Vertical} & Standing & Average & Average & Open & Close & Average &  Average \\
\midrule
RT-1-X~\cite{brohan2022rt} & \textbf{0.820} & 0.330 & 0.550 & 0.567 & 0.317 & \textbf{0.296} & \textbf{0.891} & \textbf{0.597}  & 0.534  \\
RT-2-X~\cite{zitkovich2023rt} & 0.740 & \textbf{0.740} & \underline{0.880} & \textbf{0.787} & \textbf{0.779} & 0.157 & 0.343 & 0.250  & \underline{0.607}\\
Octo-Base~\cite{mees2024octo} & 0.210 & 0.210 & 0.090 & 0.170 & 0.042 & 0.009 & 0.444 & 0.227  & 0.169 \\
OpenVLA~\cite{kim2024openvla} & 0.270 & 0.030 & 0.190 & 0.163 & 0.462 & \underline{0.194} & 0.518 & 0.356  & 0.248 \\
OpenVLA (fine-tuned)~\cite{kim2024openvla} & 0.470 & 0.080 & 0.540 & 0.363 & 0.542 & 0.102 & 0.361 & 0.231 & 0.349 \\
\midrule
Moto & \textbf{0.820} & \underline{0.500} & \textbf{0.900} & \underline{0.740} & \underline{0.604} & 0.130 & {0.732} & \underline{0.431} & \textbf{0.614} \\
Moto w/o Motion Token & 0.600 & 0.190 & 0.740 & 0.503 & 0.554 & 0.000 & \underline{0.796} & 0.398 & 0.480 \\

\bottomrule
\end{tabular}\label{tab:simpler_results}
\end{table*}

\begin{table*}
\caption{Comparison of models adopting different pre-training techniques on CALVIN (ABC$\longrightarrow$D). Avg. Len. is a comprehensive metric indicating the average number of tasks accomplished in a row across 1,000 trial sequences. ``Static RGB'' and ``Gripper RGB'' denote the RGB images from a static camera or a gripper view, respectively. ``Proprio'' is short for the proprioceptive robot state.}
\small
    \centering
    \begin{tabular}{l c c c c c c c}
    \toprule
    \multirow{2}{*}{Model} & \multirow{2}{*}{\makecell[c]{Observation Space}} & \multicolumn{6}{c}{Tasks competed in a row (1000 chains)} \\
    \cmidrule{3-8}
    & & 1 & 2 & 3 & 4 & 5 & Avg. Len. \\
    \midrule
    SuSIE~\cite{black2024zeroshot}	& Static RGB & 0.870	& 0.690	& 0.490	& 0.380	& 0.260	& 2.69 \\
    RoboFlamingo~\cite{li2024visionlanguage}	& Static RGB + Gripper RGB	 & 0.824	& 0.619	& 0.466	& 0.331	& 0.235	& 2.47 \\
    MT-R3M~\cite{wu2024unleashing} & Static RGB + Gripper RGB + Proprio & 0.529 & 0.234 & 0.105 & 0.043 & 0.018 & 0.93 \\
    GR-1~\cite{wu2024unleashing}	& Static RGB + Gripper RGB + Proprio	& 0.854	& 0.712	& {0.596}	& \textbf{0.497}	& \textbf{0.401}	& {3.06} \\
    \midrule
    Moto & Static RGB & \textbf{0.897} & \textbf{0.729} & \textbf{0.601} & 0.484 & 0.386 & \textbf{3.10} \\ 
    Moto w/o Motion Token & Static RGB & {0.779} & {0.555}& 0.380 & 0.256 & 0.167 & 2.14 \\
    \bottomrule
    \end{tabular}\label{tab:calvin_results}
\end{table*}

\noindent\textbf{Overall Performance.} 
After fine-tuning, Moto-GPT\footnote{Moto-GPT is referred to as Moto in the following tables and figures.} was evaluated on the SIMPLER and CALVIN benchmarks, demonstrating promising results as shown in Tables~\ref{tab:simpler_results} and \ref{tab:calvin_results}. It significantly outperforms Moto w/o Motion Token, which is trained from scratch without latent motion tokens, underscoring the effectiveness of transferring motion priors learned from videos to enhance robot manipulation tasks. Despite having only 98M parameters for the GPT backbone and no access to action labels during pre-training, Moto-GPT performs comparably to larger models like RT-2-X (PaLI-X 55B) and OpenVLA (Prismatic-7B) on SIMPLER. It also maintains competitiveness against OpenVLA (fine-tuned), 
which is further fine-tuned specially on the RT-1 Robot-Action trajectories, despite its pre-training data already containing action labels from this dataset.
Moto-GPT also generalizes well in the unseen CALVIN environment, outperforming baseline models that use various pre-training strategies (see supplementary material for detailed descriptions about baselines). Notably, Moto-GPT relies solely on RGB images from a static camera, achieving competitive results compared to GR-1, which is pre-trained to predict future pixel values and uses gripper RGB and proprioceptive states as additional inputs. Our findings suggest that focusing on motion dynamics rather than frame-level details is a more effective approach for learning from videos.

\begin{figure}[!t]
    \centering
    \vspace{-5pt}
    \includegraphics[width=0.42\textwidth]{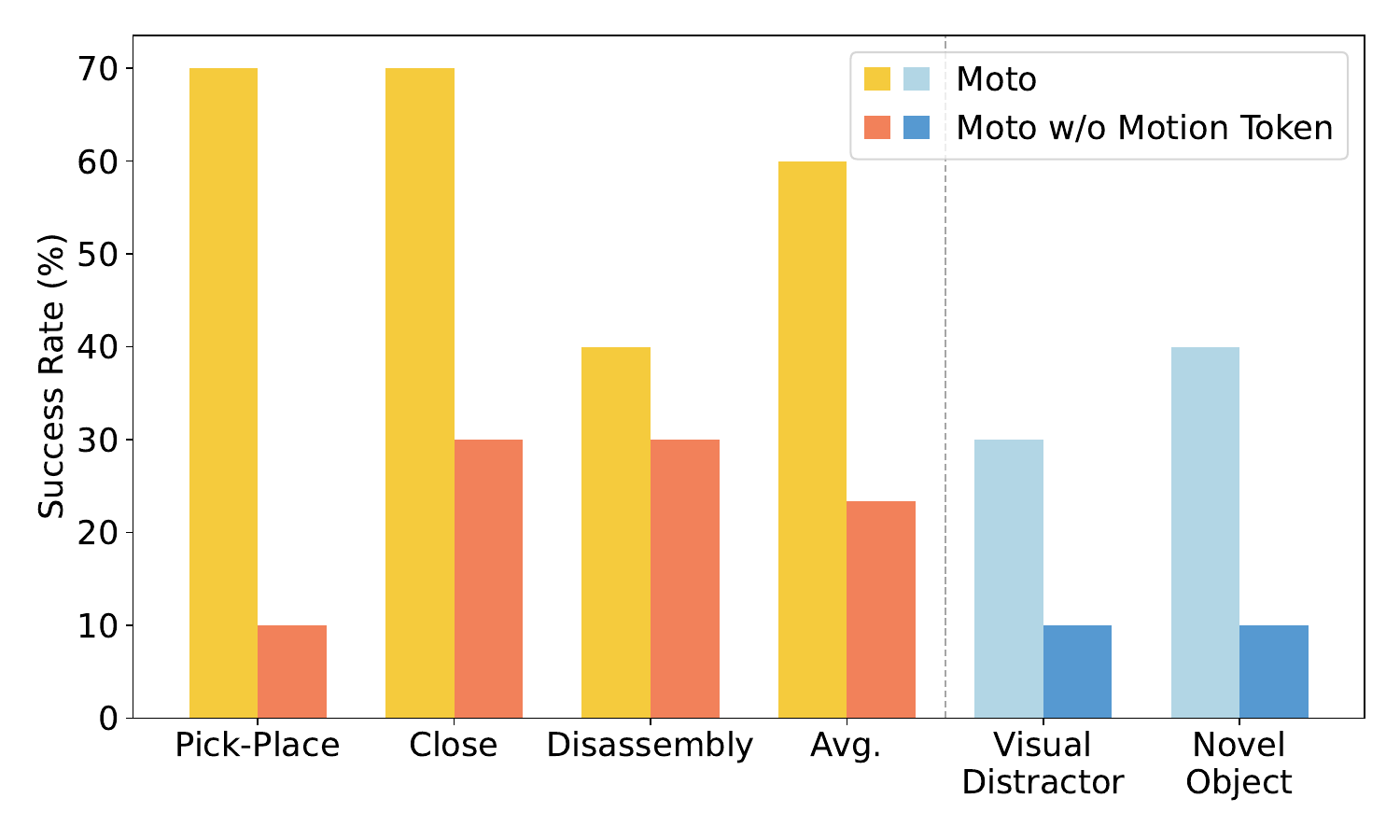}
    \vspace{-10pt}
    \caption{Evaluation results in the real-world environment.}
    \vspace{-15pt}
    \label{fig:real_robot_results}
\end{figure}

\noindent \textbf{Real-world Experiments.} 
We additionally test Moto-GPT on three real-world tasks. As shown in Fig.~\ref{fig:real_robot_results}, Moto-GPT consistently outperforms Moto w/o Motion Token on these tasks, improving the average success rate from 23.33\% to 60\%. For generalizability evaluation, we add visual distractors (Visual Distractor) or change the appearance of manipulated objects (Novel Object). Moto-GPT still enhances the average performance by 20\% and 30\% under varying conditions. This further demonstrates the robustness of Moto-GPT in real-world deployment.

\begin{figure}[!t]
    \centering
    \vspace{-5pt}\includegraphics[width=0.4\textwidth]{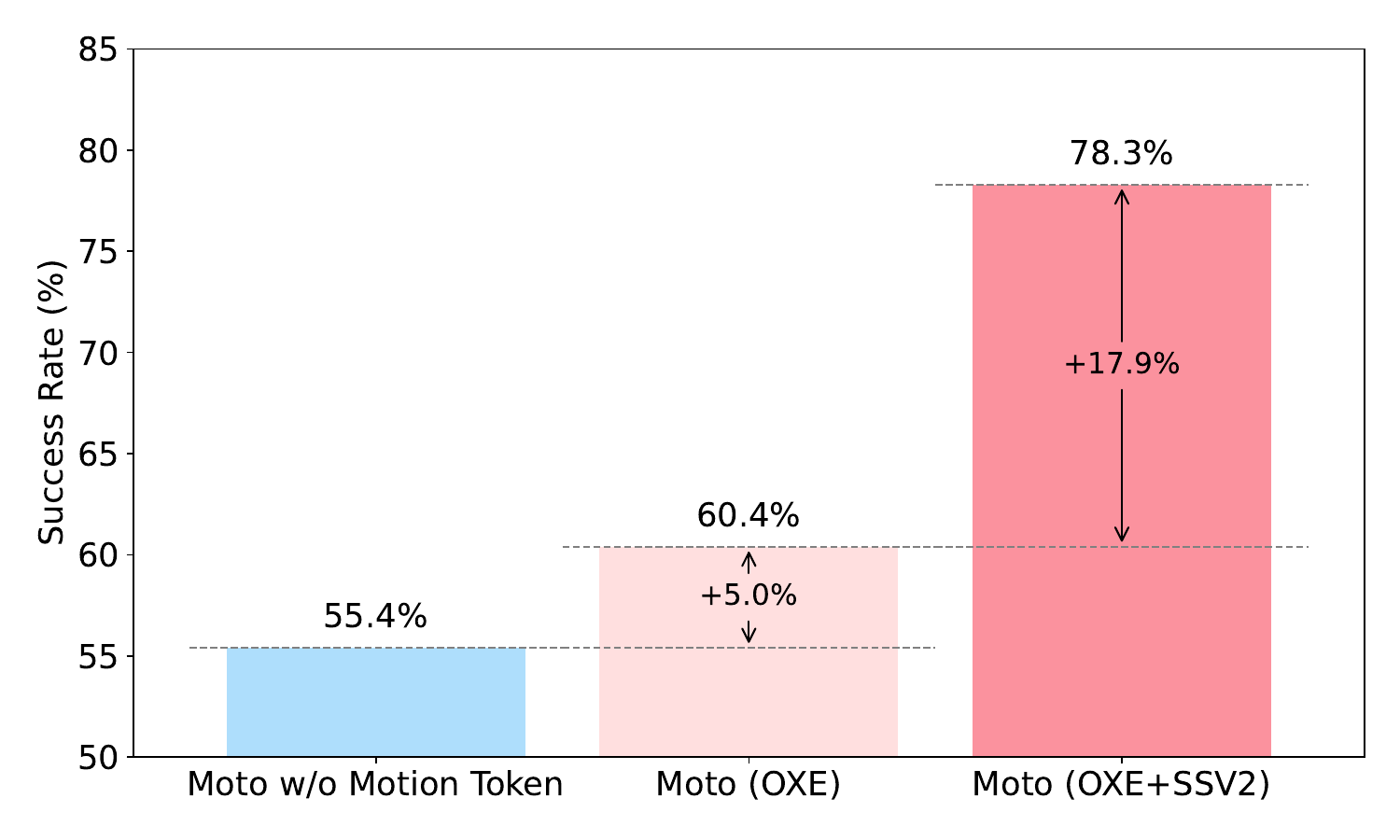}
    \vspace{-5pt}
    \caption{With additional human video (SSV2) pre-training, Moto (OXE+SSV2) significantly outperforms both Moto w/o Motion Token and Moto (OXE) on the Move Near task in SIMPLER.}
    \vspace{-10pt}
    \label{fig:move_near_human_video}
\end{figure}

\begin{figure*}[!t]
    \centering
    \includegraphics[width=1.0\textwidth]{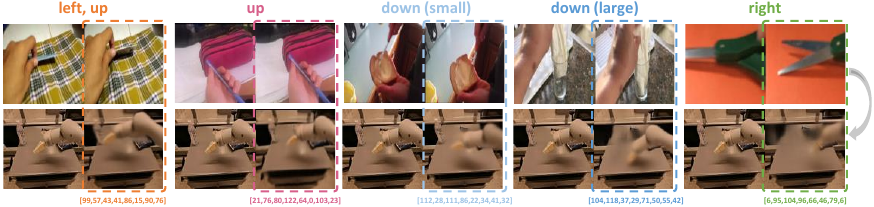}
    \vspace{-18pt}
    \caption{Visualization of motion transfer from human to robot videos via latent motion tokens.}
    \vspace{-7pt}\label{fig:human_to_robot_motion_transfer_comprehensive}
\end{figure*}

\noindent \textbf{Learning from Human Videos.}
We expand our approach to Internet-scale human videos, demonstrating their potential for large-scale robotics pre-training. Using SSV2~\cite{goyal2017something} capturing simple human hand movements, we filter out less motion-relevant videos, retaining 105k human videos for a preliminary study. These are combined with 109k OXE robot videos to train the Latent Motion Tokenizer and pre-train Moto-GPT. Fig.~\ref{fig:move_near_human_video} shows that adding human videos further enhances Moto-GPT's performance compared with Moto (OXE) which is only pre-trained on OXE videos. Additionally, latent motion tokens can act as a unified ``language''s to translate human motions into semantically aligned robot actions (Fig.~\ref{fig:human_to_robot_motion_transfer_comprehensive}). This enables not only pre-training with human videos but also in-context robot learning guided by online human demonstrations. Future work will improve model architectures and incorporate more diverse human videos to tackle complex manipulation tasks.

\begin{figure}[!t]
    \centering
    \vspace{-5pt}
    \includegraphics[width=0.38\textwidth]{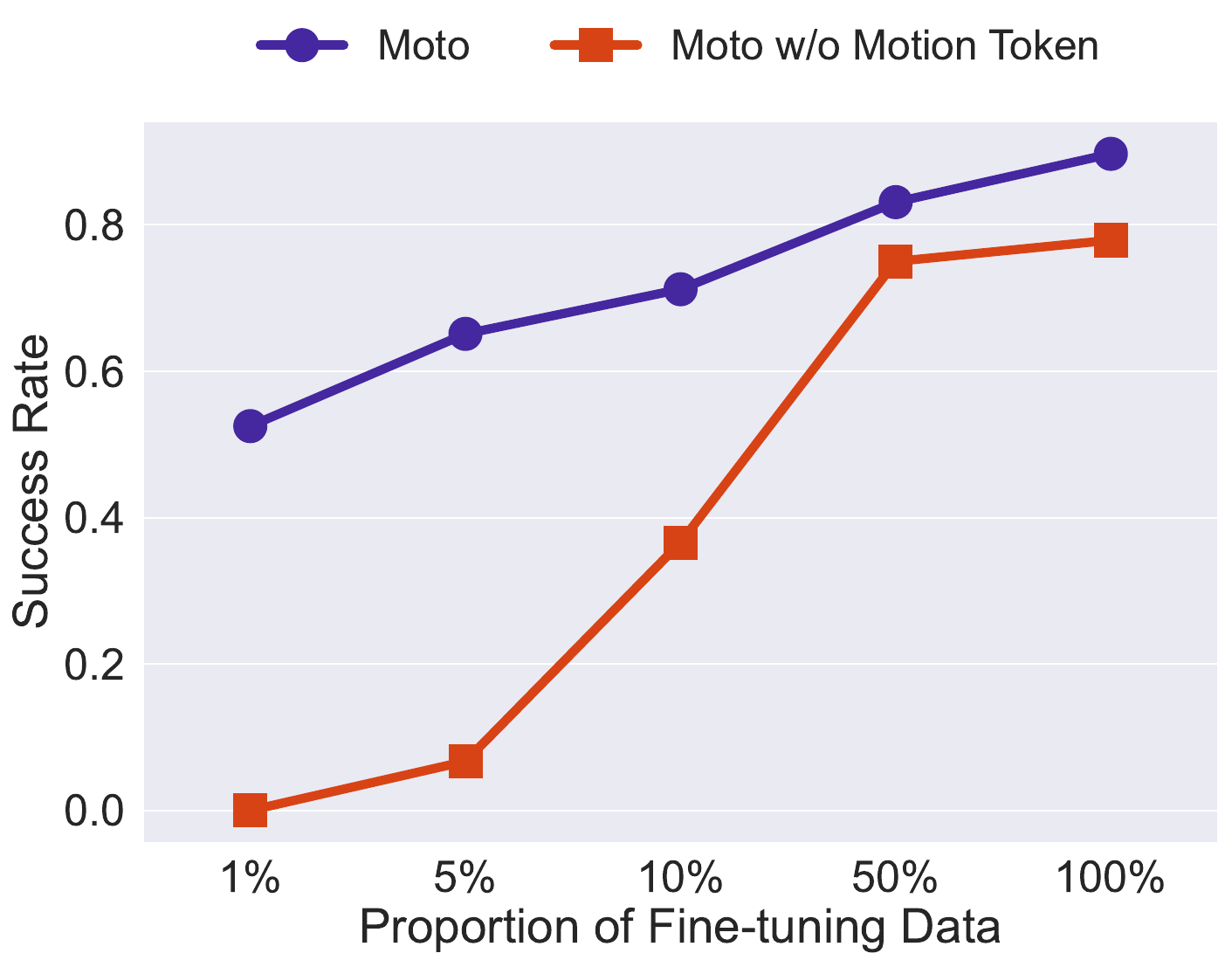}
    \caption{Task success rate of models fine-tuned with different proportions of data on CALVIN (ABC$\longrightarrow$D).}
    \vspace{-7pt}\label{fig:data_efficiency}
    \vspace{-5pt}
\end{figure}

\noindent\textbf{Data Efficiency.} 
Moto-GPT is pre-trained exclusively on videos, bypassing the need for supervised data with action labels. 
This enables pre-training on large, easily accessible video datasets, followed by fine-tuning with smaller action-labeled trajectories for policy adaptation. 
Fig.~\ref{fig:data_efficiency} shows that Moto-GPT fine-tuned with varying amounts of labeled data consistently outperforms its variant trained from scratch without latent motion tokens, especially with limited data. For instance, Moto-GPT achieves a 52.5\% success rate with just 1\% of labeled data, compared to 0\% for the variant. This demonstrates Moto-GPT's efficiency in action adaptation and its potential to improve robot manipulation tasks through large-scale video pre-training.

\begin{figure}[!t]
    \centering
    \vspace{-5pt}
    \includegraphics[width=0.38\textwidth]{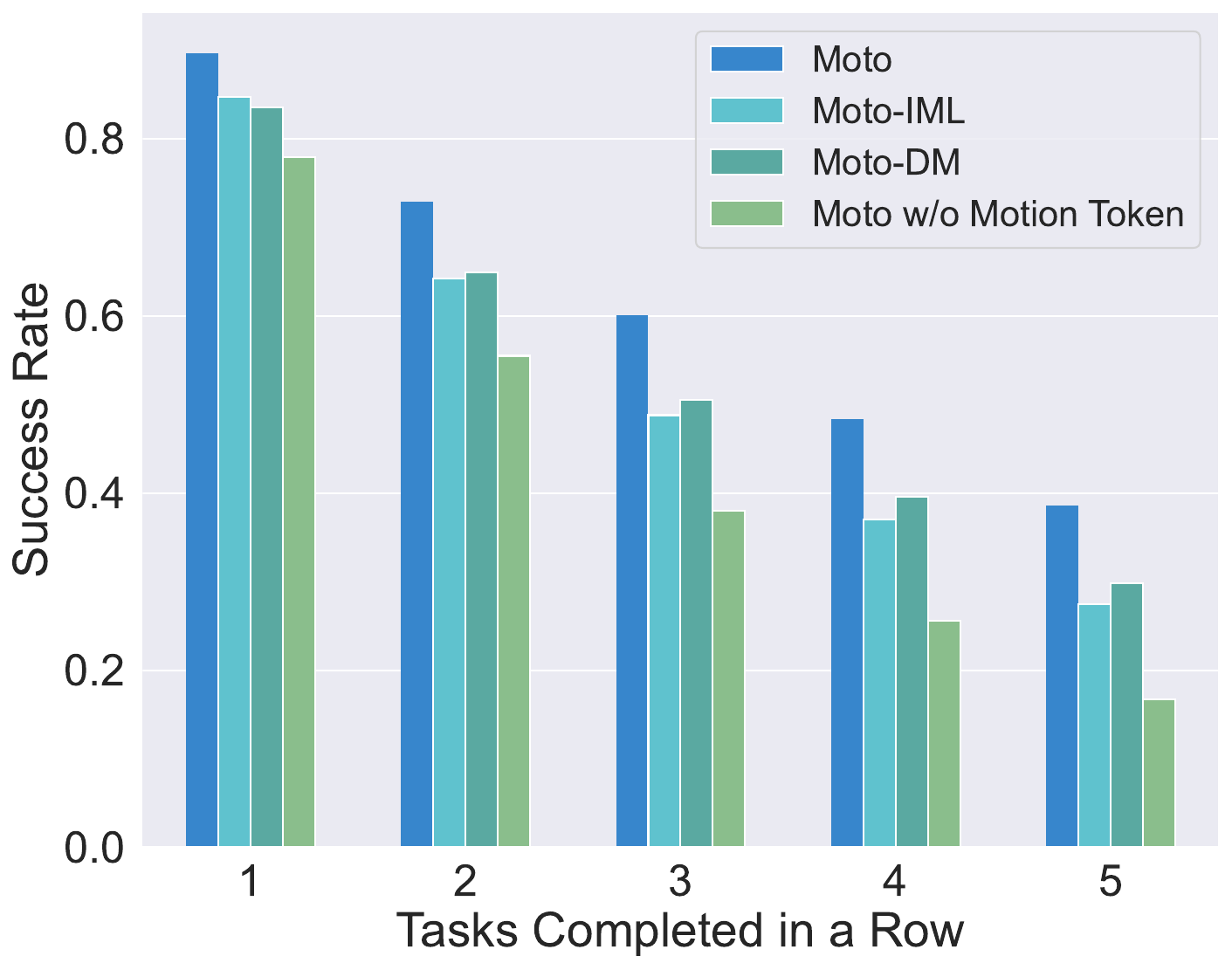}
    \caption{Ablations of Moto-GPT on CALVIN (ABC$\longrightarrow$D).}
    \label{fig:model_ablation}
        \vspace{-5pt}
\end{figure}

\noindent\textbf{Ablations on Policy Fine-tuning Methods.}
In Fig.~\ref{fig:model_ablation}, we assess Moto's co-fine-tuning strategy. Moto-IML and Moto-DM use the same pre-training as Moto-GPT but differ in fine-tuning: Moto-IML excludes the loss term for latent motion token prediction, while Moto-DM omits latent motion tokens from the input entirely. Compared to Moto w/o Motion Tokens, which is trained from scratch, both Moto-IML and Moto-DM show improved performance due to pre-training motion priors, yet they still lag behind Moto-GPT. This highlights the importance of retaining latent motion tokens in the sequence, allowing action query tokens to transfer knowledge through direct attention. Furthermore, co-fine-tuning for latent motion token prediction helps preserve the learned motion priors in Moto-GPT.

\section{Conclusion and Discussion}
We present Moto, a method that utilizes latent motion tokens as a ``language'' interface to bridge generative pre-training on video data with precise robot control. By learning motion-related priors from videos without the need for hardware-specific action labels, Moto effectively translates learned motions into actual robot actions. 
We demonstrate that Moto can not only be used for pre-training on robot videos, but also be extended to Internet-scale human videos and enable cross-embodiment knowledge transfer. 
Future research should aim to scale up the pre-training dataset and optimize fine-tuning to enhance performance across a broader range of robotic applications.

\clearpage
\section*{Acknowledgement}
The research work described in this paper was conducted in the JC STEM Lab of Autonomous Intelligent Systems funded by The Hong Kong Jockey Club Charities Trust.

{
    \small
    \bibliographystyle{ieeenat_fullname}
    \bibliography{main}

\begin{thebibliography}{62}
\providecommand{\natexlab}[1]{#1}
\providecommand{\url}[1]{\texttt{#1}}
\expandafter\ifx\csname urlstyle\endcsname\relax
  \providecommand{\doi}[1]{doi: #1}\else
  \providecommand{\doi}{doi: \begingroup \urlstyle{rm}\Url}\fi

\bibitem[Bai et~al.(2024)Bai, Geng, Mangalam, Bar, Yuille, Darrell, Malik, and Efros]{bai2024sequential}
Yutong Bai, Xinyang Geng, Karttikeya Mangalam, Amir Bar, Alan~L Yuille, Trevor Darrell, Jitendra Malik, and Alexei~A Efros.
\newblock Sequential modeling enables scalable learning for large vision models.
\newblock In \emph{Proceedings of the IEEE/CVF Conference on Computer Vision and Pattern Recognition}, pages 22861--22872, 2024.

\bibitem[Black et~al.(2024{\natexlab{a}})Black, Brown, Driess, Esmail, Equi, Finn, Fusai, Groom, Hausman, Ichter, et~al.]{black2024pi_0}
Kevin Black, Noah Brown, Danny Driess, Adnan Esmail, Michael Equi, Chelsea Finn, Niccolo Fusai, Lachy Groom, Karol Hausman, Brian Ichter, et~al.
\newblock $\pi_0$: A vision-language-action flow model for general robot control.
\newblock \emph{arXiv preprint arXiv:2410.24164}, 2024{\natexlab{a}}.

\bibitem[Black et~al.(2024{\natexlab{b}})Black, Nakamoto, Atreya, Walke, Finn, Kumar, and Levine]{black2024zeroshot}
Kevin Black, Mitsuhiko Nakamoto, Pranav Atreya, Homer~Rich Walke, Chelsea Finn, Aviral Kumar, and Sergey Levine.
\newblock Zero-shot robotic manipulation with pre-trained image-editing diffusion models.
\newblock In \emph{The Twelfth International Conference on Learning Representations}, 2024{\natexlab{b}}.

\bibitem[Brohan et~al.(2022)Brohan, Brown, Carbajal, Chebotar, Dabis, Finn, Gopalakrishnan, Hausman, Herzog, Hsu, et~al.]{brohan2022rt}
Anthony Brohan, Noah Brown, Justice Carbajal, Yevgen Chebotar, Joseph Dabis, Chelsea Finn, Keerthana Gopalakrishnan, Karol Hausman, Alex Herzog, Jasmine Hsu, et~al.
\newblock Rt-1: Robotics transformer for real-world control at scale.
\newblock \emph{arXiv preprint arXiv:2212.06817}, 2022.

\bibitem[Brown et~al.(2020)Brown, Mann, Ryder, Subbiah, Kaplan, Dhariwal, Neelakantan, Shyam, Sastry, Askell, Agarwal, Herbert-Voss, Krueger, Henighan, Child, Ramesh, Ziegler, Wu, Winter, Hesse, Chen, Sigler, Litwin, Gray, Chess, Clark, Berner, McCandlish, Radford, Sutskever, and Amodei]{brown2020language}
Tom Brown, Benjamin Mann, Nick Ryder, Melanie Subbiah, Jared~D Kaplan, Prafulla Dhariwal, Arvind Neelakantan, Pranav Shyam, Girish Sastry, Amanda Askell, Sandhini Agarwal, Ariel Herbert-Voss, Gretchen Krueger, Tom Henighan, Rewon Child, Aditya Ramesh, Daniel Ziegler, Jeffrey Wu, Clemens Winter, Chris Hesse, Mark Chen, Eric Sigler, Mateusz Litwin, Scott Gray, Benjamin Chess, Jack Clark, Christopher Berner, Sam McCandlish, Alec Radford, Ilya Sutskever, and Dario Amodei.
\newblock Language models are few-shot learners.
\newblock In \emph{Advances in Neural Information Processing Systems}, pages 1877--1901. Curran Associates, Inc., 2020.

\bibitem[Bruce et~al.(2024)Bruce, Dennis, Edwards, Parker-Holder, Shi, Hughes, Lai, Mavalankar, Steigerwald, Apps, et~al.]{bruce2024genie}
Jake Bruce, Michael~D Dennis, Ashley Edwards, Jack Parker-Holder, Yuge Shi, Edward Hughes, Matthew Lai, Aditi Mavalankar, Richie Steigerwald, Chris Apps, et~al.
\newblock Genie: Generative interactive environments.
\newblock In \emph{Forty-first International Conference on Machine Learning}, 2024.

\bibitem[Bu et~al.(2024)Bu, Zeng, Chen, Yang, Zhou, Yan, Luo, Cui, Ma, and Li]{bu2024closedloop}
Qingwen Bu, Jia Zeng, Li Chen, Yanchao Yang, Guyue Zhou, Junchi Yan, Ping Luo, Heming Cui, Yi Ma, and Hongyang Li.
\newblock Closed-loop visuomotor control with generative expectation for robotic manipulation.
\newblock In \emph{The Thirty-eighth Annual Conference on Neural Information Processing Systems}, 2024.

\bibitem[Cheang et~al.(2024)Cheang, Chen, Jing, Kong, Li, Li, Liu, Wu, Xu, Yang, et~al.]{cheang2024gr}
Chi-Lam Cheang, Guangzeng Chen, Ya Jing, Tao Kong, Hang Li, Yifeng Li, Yuxiao Liu, Hongtao Wu, Jiafeng Xu, Yichu Yang, et~al.
\newblock Gr-2: A generative video-language-action model with web-scale knowledge for robot manipulation.
\newblock \emph{arXiv preprint arXiv:2410.06158}, 2024.

\bibitem[Chen et~al.()Chen, Adebola, and Goldberg]{BerkeleyUR5Website}
Lawrence~Yunliang Chen, Simeon Adebola, and Ken Goldberg.
\newblock Berkeley {UR5} demonstration dataset.
\newblock https://sites.google.com/view/berkeley-ur5/home.

\bibitem[Chen et~al.(2024)Chen, Guo, He, Zhang, Zhang, Yang, Zhao, and Bian]{chen2024igor}
Xiaoyu Chen, Junliang Guo, Tianyu He, Chuheng Zhang, Pushi Zhang, Derek~Cathera Yang, Li Zhao, and Jiang Bian.
\newblock Igor: Image-goal representations are the atomic control units for foundation models in embodied ai.
\newblock \emph{arXiv preprint arXiv:2411.00785}, 2024.

\bibitem[Chi et~al.(2023)Chi, Xu, Feng, Cousineau, Du, Burchfiel, Tedrake, and Song]{chi2023diffusion}
Cheng Chi, Zhenjia Xu, Siyuan Feng, Eric Cousineau, Yilun Du, Benjamin Burchfiel, Russ Tedrake, and Shuran Song.
\newblock Diffusion policy: Visuomotor policy learning via action diffusion.
\newblock \emph{The International Journal of Robotics Research}, page 02783649241273668, 2023.

\bibitem[Cui et~al.(2025)Cui, Pan, Iyer, Haldar, and Pinto]{cui2025dynamo}
Zichen Cui, Hengkai Pan, Aadhithya Iyer, Siddhant Haldar, and Lerrel Pinto.
\newblock Dynamo: In-domain dynamics pretraining for visuo-motor control.
\newblock \emph{Advances in Neural Information Processing Systems}, 37:\penalty0 33933--33961, 2025.

\bibitem[Dass et~al.(2023)Dass, Yapeter, Zhang, Zhang, Pertsch, Nikolaidis, and Lim]{dass2023jacoplay}
Shivin Dass, Jullian Yapeter, Jesse Zhang, Jiahui Zhang, Karl Pertsch, Stefanos Nikolaidis, and Joseph~J. Lim.
\newblock Clvr jaco play dataset, 2023.

\bibitem[Doshi et~al.(2024)Doshi, Walke, Mees, Dasari, and Levine]{doshi2024scaling}
Ria Doshi, Homer~Rich Walke, Oier Mees, Sudeep Dasari, and Sergey Levine.
\newblock Scaling cross-embodied learning: One policy for manipulation, navigation, locomotion and aviation.
\newblock In \emph{8th Annual Conference on Robot Learning}, 2024.

\bibitem[Driess et~al.(2023)Driess, Xia, Sajjadi, Lynch, Chowdhery, Ichter, Wahid, Tompson, Vuong, Yu, et~al.]{driess2023palm}
Danny Driess, Fei Xia, Mehdi~SM Sajjadi, Corey Lynch, Aakanksha Chowdhery, Brian Ichter, Ayzaan Wahid, Jonathan Tompson, Quan Vuong, Tianhe Yu, et~al.
\newblock Palm-e: An embodied multimodal language model.
\newblock In \emph{International Conference on Machine Learning}, pages 8469--8488. PMLR, 2023.

\bibitem[Du et~al.(2024)Du, Yang, Dai, Dai, Nachum, Tenenbaum, Schuurmans, and Abbeel]{du2024learning}
Yilun Du, Sherry Yang, Bo Dai, Hanjun Dai, Ofir Nachum, Josh Tenenbaum, Dale Schuurmans, and Pieter Abbeel.
\newblock Learning universal policies via text-guided video generation.
\newblock \emph{Advances in Neural Information Processing Systems}, 36, 2024.

\bibitem[Dubey et~al.(2024)Dubey, Jauhri, Pandey, Kadian, Al-Dahle, Letman, Mathur, Schelten, Yang, Fan, et~al.]{dubey2024llama}
Abhimanyu Dubey, Abhinav Jauhri, Abhinav Pandey, Abhishek Kadian, Ahmad Al-Dahle, Aiesha Letman, Akhil Mathur, Alan Schelten, Amy Yang, Angela Fan, et~al.
\newblock The llama 3 herd of models.
\newblock \emph{arXiv preprint arXiv:2407.21783}, 2024.

\bibitem[Escontrela et~al.(2023)Escontrela, Adeniji, Yan, Jain, Peng, Goldberg, Lee, Hafner, and Abbeel]{escontrela2023video}
Alejandro Escontrela, Ademi Adeniji, Wilson Yan, Ajay Jain, Xue~Bin Peng, Ken Goldberg, Youngwoon Lee, Danijar Hafner, and Pieter Abbeel.
\newblock Video prediction models as rewards for reinforcement learning.
\newblock In \emph{Thirty-seventh Conference on Neural Information Processing Systems}, 2023.

\bibitem[Garg et~al.(2022)Garg, Vaidyanath, Kim, Song, and Ermon]{garg2022lisa}
Divyansh Garg, Skanda Vaidyanath, Kuno Kim, Jiaming Song, and Stefano Ermon.
\newblock Lisa: Learning interpretable skill abstractions from language.
\newblock \emph{Advances in Neural Information Processing Systems}, 35:\penalty0 21711--21724, 2022.

\bibitem[Goyal et~al.(2017)Goyal, Ebrahimi~Kahou, Michalski, Materzynska, Westphal, Kim, Haenel, Fruend, Yianilos, Mueller-Freitag, et~al.]{goyal2017something}
Raghav Goyal, Samira Ebrahimi~Kahou, Vincent Michalski, Joanna Materzynska, Susanne Westphal, Heuna Kim, Valentin Haenel, Ingo Fruend, Peter Yianilos, Moritz Mueller-Freitag, et~al.
\newblock The" something something" video database for learning and evaluating visual common sense.
\newblock In \emph{Proceedings of the IEEE international conference on computer vision}, pages 5842--5850, 2017.

\bibitem[Hafner et~al.(2020)Hafner, Lillicrap, Ba, and Norouzi]{Hafner2020Dream}
Danijar Hafner, Timothy Lillicrap, Jimmy Ba, and Mohammad Norouzi.
\newblock Dream to control: Learning behaviors by latent imagination.
\newblock In \emph{International Conference on Learning Representations}, 2020.

\bibitem[Haldar et~al.(2024)Haldar, Peng, and Pinto]{haldar2024baku}
Siddhant Haldar, Zhuoran Peng, and Lerrel Pinto.
\newblock {BAKU}: An efficient transformer for multi-task policy learning.
\newblock In \emph{The Thirty-eighth Annual Conference on Neural Information Processing Systems}, 2024.

\bibitem[He et~al.(2024)He, Bai, Pan, Zhang, Zhao, and Li]{he2024learning}
Haoran He, Chenjia Bai, Ling Pan, Weinan Zhang, Bin Zhao, and Xuelong Li.
\newblock Learning an actionable discrete diffusion policy via large-scale actionless video pre-training.
\newblock In \emph{The Thirty-eighth Annual Conference on Neural Information Processing Systems}, 2024.

\bibitem[He et~al.(2022)He, Chen, Xie, Li, Doll{\'a}r, and Girshick]{he2022masked}
Kaiming He, Xinlei Chen, Saining Xie, Yanghao Li, Piotr Doll{\'a}r, and Ross Girshick.
\newblock Masked autoencoders are scalable vision learners.
\newblock In \emph{Proceedings of the IEEE/CVF conference on computer vision and pattern recognition}, pages 16000--16009, 2022.

\bibitem[Jiang et~al.(2023)Jiang, Gupta, Zhang, Wang, Dou, Chen, Fei-Fei, Anandkumar, Zhu, and Fan]{jiang2023vima}
Yunfan Jiang, Agrim Gupta, Zichen Zhang, Guanzhi Wang, Yongqiang Dou, Yanjun Chen, Li Fei-Fei, Anima Anandkumar, Yuke Zhu, and Linxi Fan.
\newblock Vima: robot manipulation with multimodal prompts.
\newblock In \emph{Proceedings of the 40th International Conference on Machine Learning}, pages 14975--15022, 2023.

\bibitem[Ke et~al.(2024)Ke, Gkanatsios, and Fragkiadaki]{ke2024d}
Tsung-Wei Ke, Nikolaos Gkanatsios, and Katerina Fragkiadaki.
\newblock 3d diffuser actor: Policy diffusion with 3d scene representations.
\newblock In \emph{ICRA 2024 Workshop{\textemdash}Back to the Future: Robot Learning Going Probabilistic}, 2024.

\bibitem[Kim et~al.(2024)Kim, Pertsch, Karamcheti, Xiao, Balakrishna, Nair, Rafailov, Foster, Lam, Sanketi, et~al.]{kim2024openvla}
Moo~Jin Kim, Karl Pertsch, Siddharth Karamcheti, Ted Xiao, Ashwin Balakrishna, Suraj Nair, Rafael Rafailov, Ethan Foster, Grace Lam, Pannag Sanketi, et~al.
\newblock Openvla: An open-source vision-language-action model.
\newblock \emph{arXiv preprint arXiv:2406.09246}, 2024.

\bibitem[Ko et~al.(2024)Ko, Mao, Du, Sun, and Tenenbaum]{ko2024learning}
Po-Chen Ko, Jiayuan Mao, Yilun Du, Shao-Hua Sun, and Joshua~B. Tenenbaum.
\newblock Learning to act from actionless videos through dense correspondences.
\newblock In \emph{The Twelfth International Conference on Learning Representations}, 2024.

\bibitem[Li et~al.(2024{\natexlab{a}})Li, Wu, Huang, Cheang, Wang, and Kong]{li2024gr}
Peiyan Li, Hongtao Wu, Yan Huang, Chilam Cheang, Liang Wang, and Tao Kong.
\newblock Gr-mg: Leveraging partially annotated data via multi-modal goal conditioned policy.
\newblock \emph{arXiv preprint arXiv:2408.14368}, 2024{\natexlab{a}}.

\bibitem[Li et~al.(2024{\natexlab{b}})Li, Liang, Wang, Luo, Chen, Liao, Wei, Deng, Xu, Zhang, et~al.]{li2024cogact}
Qixiu Li, Yaobo Liang, Zeyu Wang, Lin Luo, Xi Chen, Mozheng Liao, Fangyun Wei, Yu Deng, Sicheng Xu, Yizhong Zhang, et~al.
\newblock Cogact: A foundational vision-language-action model for synergizing cognition and action in robotic manipulation.
\newblock \emph{arXiv preprint arXiv:2411.19650}, 2024{\natexlab{b}}.

\bibitem[Li et~al.(2024{\natexlab{c}})Li, Hsu, Gu, Pertsch, Mees, Walke, Fu, Lunawat, Sieh, Kirmani, et~al.]{li2024evaluating}
Xuanlin Li, Kyle Hsu, Jiayuan Gu, Karl Pertsch, Oier Mees, Homer~Rich Walke, Chuyuan Fu, Ishikaa Lunawat, Isabel Sieh, Sean Kirmani, et~al.
\newblock Evaluating real-world robot manipulation policies in simulation.
\newblock \emph{arXiv preprint arXiv:2405.05941}, 2024{\natexlab{c}}.

\bibitem[Li et~al.(2024{\natexlab{d}})Li, Liu, Zhang, Yu, Xu, Wu, Cheang, Jing, Zhang, Liu, Li, and Kong]{li2024visionlanguage}
Xinghang Li, Minghuan Liu, Hanbo Zhang, Cunjun Yu, Jie Xu, Hongtao Wu, Chilam Cheang, Ya Jing, Weinan Zhang, Huaping Liu, Hang Li, and Tao Kong.
\newblock Vision-language foundation models as effective robot imitators.
\newblock In \emph{The Twelfth International Conference on Learning Representations}, 2024{\natexlab{d}}.

\bibitem[Liang et~al.(2024{\natexlab{a}})Liang, Liu, Ozguroglu, Sudhakar, Dave, Tokmakov, Song, and Vondrick]{liang2024dreamitate}
Junbang Liang, Ruoshi Liu, Ege Ozguroglu, Sruthi Sudhakar, Achal Dave, Pavel Tokmakov, Shuran Song, and Carl Vondrick.
\newblock Dreamitate: Real-world visuomotor policy learning via video generation.
\newblock \emph{arXiv preprint arXiv:2406.16862}, 2024{\natexlab{a}}.

\bibitem[Liang et~al.(2024{\natexlab{b}})Liang, Mu, Ma, Tomizuka, Ding, and Luo]{liang2024skilldiffuser}
Zhixuan Liang, Yao Mu, Hengbo Ma, Masayoshi Tomizuka, Mingyu Ding, and Ping Luo.
\newblock Skilldiffuser: Interpretable hierarchical planning via skill abstractions in diffusion-based task execution.
\newblock In \emph{Proceedings of the IEEE/CVF Conference on Computer Vision and Pattern Recognition}, pages 16467--16476, 2024{\natexlab{b}}.

\bibitem[Liu et~al.(2024)Liu, Yan, Zheng, Feng, Huang, and Ma]{liu2024robouniview}
Fanfan Liu, Feng Yan, Liming Zheng, Chengjian Feng, Yiyang Huang, and Lin Ma.
\newblock Robouniview: Visual-language model with unified view representation for robotic manipulation.
\newblock \emph{arXiv preprint arXiv:2406.18977}, 2024.

\bibitem[Luo et~al.(2023)Luo, Xu, Geng, Feng, Fang, Tan, Schaal, and Levine]{luo2023multistage}
Jianlan Luo, Charles Xu, Xinyang Geng, Gilbert Feng, Kuan Fang, Liam Tan, Stefan Schaal, and Sergey Levine.
\newblock Multi-stage cable routing through hierarchical imitation learning.
\newblock \emph{arXiv pre-print}, 2023.

\bibitem[Ma et~al.(2023)Ma, Sodhani, Jayaraman, Bastani, Kumar, and Zhang]{ma2023vip}
Yecheng~Jason Ma, Shagun Sodhani, Dinesh Jayaraman, Osbert Bastani, Vikash Kumar, and Amy Zhang.
\newblock {VIP}: Towards universal visual reward and representation via value-implicit pre-training.
\newblock In \emph{The Eleventh International Conference on Learning Representations}, 2023.

\bibitem[Mandlekar et~al.(2019)Mandlekar, Booher, Spero, Tung, Gupta, Zhu, Garg, Savarese, and Fei-Fei]{mandlekar2019scaling}
Ajay Mandlekar, Jonathan Booher, Max Spero, Albert Tung, Anchit Gupta, Yuke Zhu, Animesh Garg, Silvio Savarese, and Li Fei-Fei.
\newblock Scaling robot supervision to hundreds of hours with roboturk: Robotic manipulation dataset through human reasoning and dexterity.
\newblock In \emph{2019 IEEE/RSJ International Conference on Intelligent Robots and Systems (IROS)}, pages 1048--1055. IEEE, 2019.

\bibitem[Mees et~al.(2022)Mees, Hermann, Rosete-Beas, and Burgard]{mees2022calvin}
Oier Mees, Lukas Hermann, Erick Rosete-Beas, and Wolfram Burgard.
\newblock Calvin: A benchmark for language-conditioned policy learning for long-horizon robot manipulation tasks.
\newblock \emph{IEEE Robotics and Automation Letters}, 7\penalty0 (3):\penalty0 7327--7334, 2022.

\bibitem[Mees et~al.(2023)Mees, Borja-Diaz, and Burgard]{mees23hulc2}
Oier Mees, Jessica Borja-Diaz, and Wolfram Burgard.
\newblock Grounding language with visual affordances over unstructured data.
\newblock In \emph{Proceedings of the IEEE International Conference on Robotics and Automation (ICRA)}, London, UK, 2023.

\bibitem[Mees et~al.(2024)Mees, Ghosh, Pertsch, Black, Walke, Dasari, Hejna, Kreiman, Xu, Luo, Tan, Sadigh, Finn, and Levine]{mees2024octo}
Oier Mees, Dibya Ghosh, Karl Pertsch, Kevin Black, Homer~Rich Walke, Sudeep Dasari, Joey Hejna, Tobias Kreiman, Charles Xu, Jianlan Luo, You~Liang Tan, Dorsa Sadigh, Chelsea Finn, and Sergey Levine.
\newblock Octo: An open-source generalist robot policy.
\newblock In \emph{First Workshop on Vision-Language Models for Navigation and Manipulation at ICRA 2024}, 2024.

\bibitem[Nair et~al.(2023)Nair, Rajeswaran, Kumar, Finn, and Gupta]{nair2023r3m}
Suraj Nair, Aravind Rajeswaran, Vikash Kumar, Chelsea Finn, and Abhinav Gupta.
\newblock R3m: A universal visual representation for robot manipulation.
\newblock In \emph{Conference on Robot Learning}, pages 892--909. PMLR, 2023.

\bibitem[Ouyang et~al.(2022)Ouyang, Wu, Jiang, Almeida, Wainwright, Mishkin, Zhang, Agarwal, Slama, Ray, et~al.]{ouyang2022training}
Long Ouyang, Jeffrey Wu, Xu Jiang, Diogo Almeida, Carroll Wainwright, Pamela Mishkin, Chong Zhang, Sandhini Agarwal, Katarina Slama, Alex Ray, et~al.
\newblock Training language models to follow instructions with human feedback.
\newblock \emph{Advances in neural information processing systems}, 35:\penalty0 27730--27744, 2022.

\bibitem[Pari et~al.(2021)Pari, Shafiullah, Arunachalam, and Pinto]{pari2021surprising}
Jyothish Pari, Nur~Muhammad Shafiullah, Sridhar~Pandian Arunachalam, and Lerrel Pinto.
\newblock The surprising effectiveness of representation learning for visual imitation, 2021.

\bibitem[Radford(2018)]{radford2018improving}
Alec Radford.
\newblock Improving language understanding by generative pre-training.
\newblock 2018.

\bibitem[Raffel et~al.(2020)Raffel, Shazeer, Roberts, Lee, Narang, Matena, Zhou, Li, and Liu]{raffel2020exploring}
Colin Raffel, Noam Shazeer, Adam Roberts, Katherine Lee, Sharan Narang, Michael Matena, Yanqi Zhou, Wei Li, and Peter~J Liu.
\newblock Exploring the limits of transfer learning with a unified text-to-text transformer.
\newblock \emph{Journal of machine learning research}, 21\penalty0 (140):\penalty0 1--67, 2020.

\bibitem[Reed et~al.(2022)Reed, Zolna, Parisotto, Colmenarejo, Novikov, Barth-maron, Gim{\'e}nez, Sulsky, Kay, Springenberg, Eccles, Bruce, Razavi, Edwards, Heess, Chen, Hadsell, Vinyals, Bordbar, and de~Freitas]{reed2022a}
Scott Reed, Konrad Zolna, Emilio Parisotto, Sergio~G{\'o}mez Colmenarejo, Alexander Novikov, Gabriel Barth-maron, Mai Gim{\'e}nez, Yury Sulsky, Jackie Kay, Jost~Tobias Springenberg, Tom Eccles, Jake Bruce, Ali Razavi, Ashley Edwards, Nicolas Heess, Yutian Chen, Raia Hadsell, Oriol Vinyals, Mahyar Bordbar, and Nando de Freitas.
\newblock A generalist agent.
\newblock \emph{Transactions on Machine Learning Research}, 2022.
\newblock Featured Certification, Outstanding Certification.

\bibitem[Rosete-Beas et~al.(2022)Rosete-Beas, Mees, Kalweit, Boedecker, and Burgard]{rosete2022tacorl}
Erick Rosete-Beas, Oier Mees, Gabriel Kalweit, Joschka Boedecker, and Wolfram Burgard.
\newblock Latent plans for task agnostic offline reinforcement learning.
\newblock 2022.

\bibitem[Schmidt and Jiang(2023)]{schmidt2023learning}
Dominik Schmidt and Minqi Jiang.
\newblock Learning to act without actions.
\newblock \emph{arXiv preprint arXiv:2312.10812}, 2023.

\bibitem[Touvron et~al.(2023)Touvron, Lavril, Izacard, Martinet, Lachaux, Lacroix, Rozi{\`e}re, Goyal, Hambro, Azhar, et~al.]{touvron2023llama}
Hugo Touvron, Thibaut Lavril, Gautier Izacard, Xavier Martinet, Marie-Anne Lachaux, Timoth{\'e}e Lacroix, Baptiste Rozi{\`e}re, Naman Goyal, Eric Hambro, Faisal Azhar, et~al.
\newblock Llama: Open and efficient foundation language models.
\newblock \emph{arXiv preprint arXiv:2302.13971}, 2023.

\bibitem[Van Den~Oord et~al.(2017)Van Den~Oord, Vinyals, et~al.]{van2017neural}
Aaron Van Den~Oord, Oriol Vinyals, et~al.
\newblock Neural discrete representation learning.
\newblock \emph{Advances in neural information processing systems}, 30, 2017.

\bibitem[Vuong et~al.(2023)Vuong, Levine, Walke, Pertsch, Singh, Doshi, Xu, Luo, Tan, Shah, Finn, Du, Kim, Khazatsky, Yang, Zhao, Goldberg, Hoque, Chen, Adebola, Sukhatme, Salhotra, Dass, Pinto, Cui, Haldar, Rai, Shafiullah, Zhu, Zhu, Nasiriany, Song, Chi, Pan, Burgard, Mees, Huang, Pathak, Bahl, Mendonca, Zhou, Srirama, Dasari, Lu, Fang, Fang, Christensen, Tomizuka, Zhan, Ding, Xu, Zhu, Tian, Lee, Sadigh, Cui, Belkhale, Sundaresan, Darrell, Malik, Radosavovic, Bohg, Srinivasan, Wang, Hansen, Wu, Yan, Su, Gu, Li, Suenderhauf, Rana, Burgess-Limerick, Ceola, Kawaharazuka, Kanazawa, Matsushima, Matsuo, Iwasawa, Furuta, Oh, Harada, Osa, Tang, Kroemer, Sharma, Zhang, Kim, Cho, Han, Kim, Lim, Johns, Palo, Stulp, Raffin, Bustamante, Silv{\'e}rio, Padalkar, Peters, Sch{\"o}lkopf, B{\"u}chler, Schneider, Guist, Wu, Tian, Shi, Li, Wang, Zhang, Amor, Zhou, Majd, Ott, Schiavi, Mart{\'\i}n-Mart{\'\i}n, Shah, Bisk, Bingham, Yu, Jain, Xiao, Hausman, Chan, Herzog, Xu, Kirmani, Vanhoucke, Julian, Lee, Ding, Chebotar, Tan,
  Liang, Mordatch, Rao, Lu, Gopalakrishnan, Welker, Joshi, Devin, Irpan, Moore, Wahid, Wu, Chen, Wohlhart, Bewley, Zhou, Leal, Kalashnikov, Sanketi, Fu, Xu, Xu, brian ichter, Hsu, Xu, Brohan, Sermanet, Heess, Ahn, Rafailov, Pooley, Byrne, Davchev, Oslund, Schaal, Jain, Go, Xia, Tompson, Armstrong, and Driess]{vuong2023open}
Quan Vuong, Sergey Levine, Homer~Rich Walke, Karl Pertsch, Anikait Singh, Ria Doshi, Charles Xu, Jianlan Luo, Liam Tan, Dhruv Shah, Chelsea Finn, Max Du, Moo~Jin Kim, Alexander Khazatsky, Jonathan~Heewon Yang, Tony~Z. Zhao, Ken Goldberg, Ryan Hoque, Lawrence~Yunliang Chen, Simeon Adebola, Gaurav~S. Sukhatme, Gautam Salhotra, Shivin Dass, Lerrel Pinto, Zichen~Jeff Cui, Siddhant Haldar, Anant Rai, Nur Muhammad~Mahi Shafiullah, Yuke Zhu, Yifeng Zhu, Soroush Nasiriany, Shuran Song, Cheng Chi, Chuer Pan, Wolfram Burgard, Oier Mees, Chenguang Huang, Deepak Pathak, Shikhar Bahl, Russell Mendonca, Gaoyue Zhou, Mohan~Kumar Srirama, Sudeep Dasari, Cewu Lu, Hao-Shu Fang, Hongjie Fang, Henrik~I Christensen, Masayoshi Tomizuka, Wei Zhan, Mingyu Ding, Chenfeng Xu, Xinghao Zhu, Ran Tian, Youngwoon Lee, Dorsa Sadigh, Yuchen Cui, Suneel Belkhale, Priya Sundaresan, Trevor Darrell, Jitendra Malik, Ilija Radosavovic, Jeannette Bohg, Krishnan Srinivasan, Xiaolong Wang, Nicklas Hansen, Yueh-Hua Wu, Ge Yan, Hao Su, Jiayuan Gu,
  Xuanlin Li, Niko Suenderhauf, Krishan Rana, Ben Burgess-Limerick, Federico Ceola, Kento Kawaharazuka, Naoaki Kanazawa, Tatsuya Matsushima, Yutaka Matsuo, Yusuke Iwasawa, Hiroki Furuta, Jihoon Oh, Tatsuya Harada, Takayuki Osa, Yujin Tang, Oliver Kroemer, Mohit Sharma, Kevin~Lee Zhang, Beomjoon Kim, Yoonyoung Cho, Junhyek Han, Jaehyung Kim, Joseph~J Lim, Edward Johns, Norman~Di Palo, Freek Stulp, Antonin Raffin, Samuel Bustamante, Jo{\~a}o Silv{\'e}rio, Abhishek Padalkar, Jan Peters, Bernhard Sch{\"o}lkopf, Dieter B{\"u}chler, Jan Schneider, Simon Guist, Jiajun Wu, Stephen Tian, Haochen Shi, Yunzhu Li, Yixuan Wang, Mingtong Zhang, Heni~Ben Amor, Yifan Zhou, Keyvan Majd, Lionel Ott, Giulio Schiavi, Roberto Mart{\'\i}n-Mart{\'\i}n, Rutav Shah, Yonatan Bisk, Jeffrey~T Bingham, Tianhe Yu, Vidhi Jain, Ted Xiao, Karol Hausman, Christine Chan, Alexander Herzog, Zhuo Xu, Sean Kirmani, Vincent Vanhoucke, Ryan Julian, Lisa Lee, Tianli Ding, Yevgen Chebotar, Jie Tan, Jacky Liang, Igor Mordatch, Kanishka Rao, Yao Lu,
  Keerthana Gopalakrishnan, Stefan Welker, Nikhil~J Joshi, Coline~Manon Devin, Alex Irpan, Sherry Moore, Ayzaan Wahid, Jialin Wu, Xi Chen, Paul Wohlhart, Alex Bewley, Wenxuan Zhou, Isabel Leal, Dmitry Kalashnikov, Pannag~R Sanketi, Chuyuan Fu, Ying Xu, Sichun Xu, brian ichter, Jasmine Hsu, Peng Xu, Anthony Brohan, Pierre Sermanet, Nicolas Heess, Michael Ahn, Rafael Rafailov, Acorn Pooley, Kendra Byrne, Todor Davchev, Kenneth Oslund, Stefan Schaal, Ajinkya Jain, Keegan Go, Fei Xia, Jonathan Tompson, Travis Armstrong, and Danny Driess.
\newblock Open x-embodiment: Robotic learning datasets and {RT}-x models.
\newblock In \emph{Towards Generalist Robots: Learning Paradigms for Scalable Skill Acquisition @ CoRL2023}, 2023.

\bibitem[Walke et~al.(2023)Walke, Black, Lee, Kim, Du, Zheng, Zhao, Hansen-Estruch, Vuong, He, Myers, Fang, Finn, and Levine]{walke2023bridgedata}
Homer Walke, Kevin Black, Abraham Lee, Moo~Jin Kim, Max Du, Chongyi Zheng, Tony Zhao, Philippe Hansen-Estruch, Quan Vuong, Andre He, Vivek Myers, Kuan Fang, Chelsea Finn, and Sergey Levine.
\newblock Bridgedata v2: A dataset for robot learning at scale.
\newblock In \emph{Conference on Robot Learning (CoRL)}, 2023.

\bibitem[Wu et~al.(2024{\natexlab{a}})Wu, Jing, Cheang, Chen, Xu, Li, Liu, Li, and Kong]{wu2024unleashing}
Hongtao Wu, Ya Jing, Chilam Cheang, Guangzeng Chen, Jiafeng Xu, Xinghang Li, Minghuan Liu, Hang Li, and Tao Kong.
\newblock Unleashing large-scale video generative pre-training for visual robot manipulation.
\newblock In \emph{The Twelfth International Conference on Learning Representations}, 2024{\natexlab{a}}.

\bibitem[Wu et~al.(2024{\natexlab{b}})Wu, Yin, Feng, He, Li, HAO, and Long]{wu2024ivideogpt}
Jialong Wu, Shaofeng Yin, Ningya Feng, Xu He, Dong Li, Jianye HAO, and Mingsheng Long.
\newblock ivideo{GPT}: Interactive video{GPT}s are scalable world models.
\newblock In \emph{The Thirty-eighth Annual Conference on Neural Information Processing Systems}, 2024{\natexlab{b}}.

\bibitem[Yang et~al.(2024{\natexlab{a}})Yang, Du, Ghasemipour, Tompson, Kaelbling, Schuurmans, and Abbeel]{yang2024learning}
Sherry Yang, Yilun Du, Seyed Kamyar~Seyed Ghasemipour, Jonathan Tompson, Leslie~Pack Kaelbling, Dale Schuurmans, and Pieter Abbeel.
\newblock Learning interactive real-world simulators.
\newblock In \emph{The Twelfth International Conference on Learning Representations}, 2024{\natexlab{a}}.

\bibitem[Yang et~al.(2024{\natexlab{b}})Yang, Walker, Parker-Holder, Du, Bruce, Barreto, Abbeel, and Schuurmans]{yang2024video}
Sherry Yang, Jacob Walker, Jack Parker-Holder, Yilun Du, Jake Bruce, Andre Barreto, Pieter Abbeel, and Dale Schuurmans.
\newblock Video as the new language for real-world decision making.
\newblock \emph{arXiv preprint arXiv:2402.17139}, 2024{\natexlab{b}}.

\bibitem[Ye et~al.(2024)Ye, Jang, Jeon, Joo, Yang, Peng, Mandlekar, Tan, Chao, Lin, et~al.]{ye2024latent}
Seonghyeon Ye, Joel Jang, Byeongguk Jeon, Sejune Joo, Jianwei Yang, Baolin Peng, Ajay Mandlekar, Reuben Tan, Yu-Wei Chao, Bill~Yuchen Lin, et~al.
\newblock Latent action pretraining from videos.
\newblock \emph{arXiv preprint arXiv:2410.11758}, 2024.

\bibitem[Zhen et~al.(2024)Zhen, Qiu, Chen, Yang, Yan, Du, Hong, and Gan]{zhen20243d}
Haoyu Zhen, Xiaowen Qiu, Peihao Chen, Jincheng Yang, Xin Yan, Yilun Du, Yining Hong, and Chuang Gan.
\newblock 3d-vla: A 3d vision-language-action generative world model.
\newblock \emph{arXiv preprint arXiv:2403.09631}, 2024.

\bibitem[Zhou et~al.(2023)Zhou, Dean, Srirama, Rajeswaran, Pari, Hatch, Jain, Yu, Abbeel, Pinto, Finn, and Gupta]{zhou2023train}
Gaoyue Zhou, Victoria Dean, Mohan~Kumar Srirama, Aravind Rajeswaran, Jyothish Pari, Kyle Hatch, Aryan Jain, Tianhe Yu, Pieter Abbeel, Lerrel Pinto, Chelsea Finn, and Abhinav Gupta.
\newblock Train offline, test online: A real robot learning benchmark.
\newblock In \emph{2023 IEEE International Conference on Robotics and Automation (ICRA)}, 2023.

\bibitem[Zhu et~al.(2022)Zhu, Joshi, Stone, and Zhu]{zhu2022viola}
Yifeng Zhu, Abhishek Joshi, Peter Stone, and Yuke Zhu.
\newblock Viola: Imitation learning for vision-based manipulation with object proposal priors.
\newblock \emph{6th Annual Conference on Robot Learning (CoRL)}, 2022.

\bibitem[Zitkovich et~al.(2023)Zitkovich, Yu, Xu, Xu, Xiao, Xia, Wu, Wohlhart, Welker, Wahid, et~al.]{zitkovich2023rt}
Brianna Zitkovich, Tianhe Yu, Sichun Xu, Peng Xu, Ted Xiao, Fei Xia, Jialin Wu, Paul Wohlhart, Stefan Welker, Ayzaan Wahid, et~al.
\newblock Rt-2: Vision-language-action models transfer web knowledge to robotic control.
\newblock In \emph{Conference on Robot Learning}, pages 2165--2183. PMLR, 2023.

\end{thebibliography}
}

\clearpage
\setcounter{page}{1}
\maketitlesupplementary

\section{Details on Experiment Setup}
\subsection{Benchmarks}

\begin{figure}[!h]
    \centering
    \includegraphics[width=0.47\textwidth]{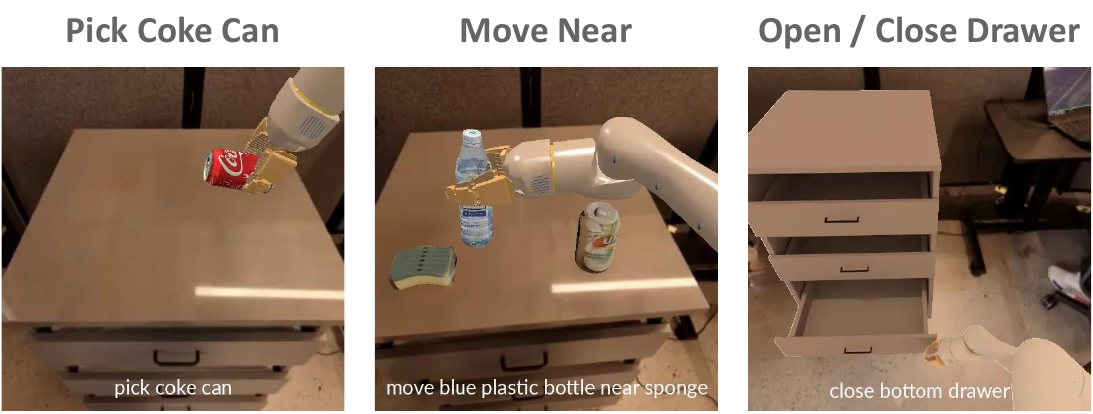}
    \caption{
    Illustration of the evaluation tasks in SIMPLER~\cite{li2024evaluating}.
    }
    \label{fig:simpler_benchmark_illustration}
\end{figure}

\paragraph{SIMPLER.} On the SIMPLER benchmark, we focus on three tasks concerning the Google Everyday Robot embodiment: \texttt{Pick Coke Can}, \texttt{Move Near}, and \texttt{Open/Close Drawer}, as illustrated in Fig.~\ref{fig:simpler_benchmark_illustration}. The ``Pick Coke Can'' task involves grasping and lifting the empty coke can in three different orientations: horizontal laying, vertical laying, and standing. The ``Move Near'' task places 3 (out of 8) objects in a triangle pattern on the tabletop and instructs the robot to move a designated source object near another object as the target. 
The ``Open/Close Drawer'' task requires the robot to open or close a specific drawer from the top, middle, or down position of a cabinet. 

\begin{figure}[!t]
    \centering
    \includegraphics[width=0.47\textwidth]{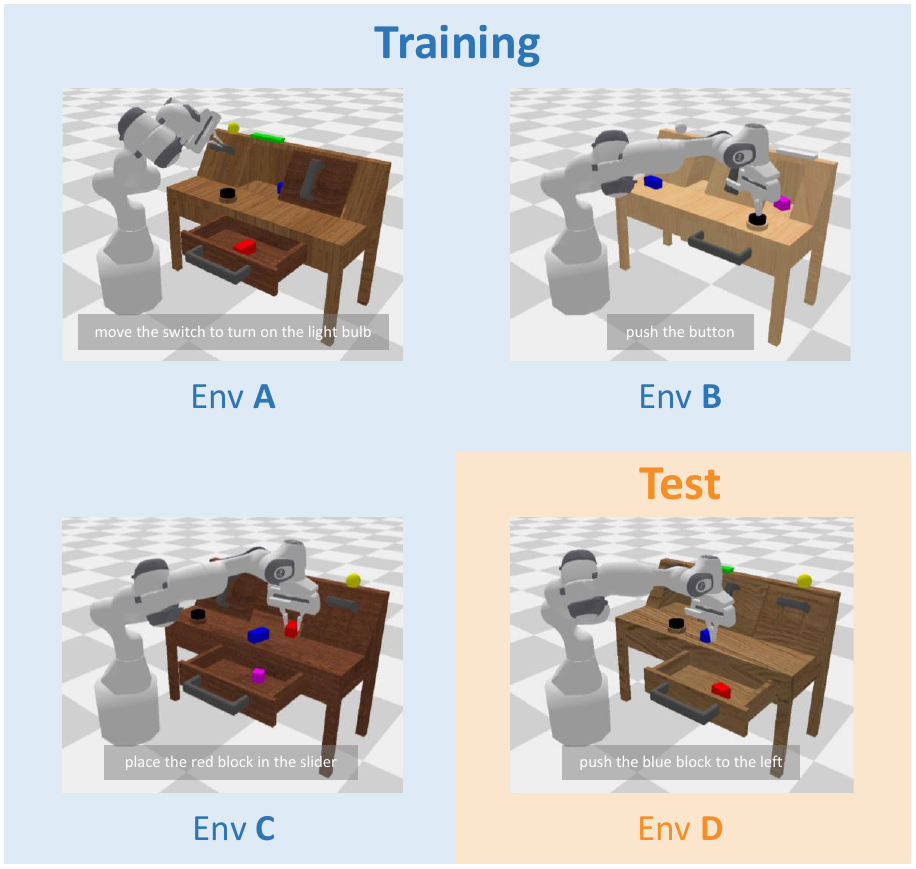}
    \caption{
    Illustration of the four different environments in CALVIN, adapted from the original figure in~\citet{mees2022calvin}.
    }
    \label{fig:calvin_benchmark_illustration}
\end{figure}

\paragraph{CALVIN (ABC$\longrightarrow$D).} The CALVIN benchmark uses a Franka Emika Panda robot embodiment. It evaluates long-horizon task completion with unconstrained language instructions. In each trial, the robot should accomplish 5 (out of 34) tasks in a row. 
There are four different environments (A, B, C, D), each containing a desk with a sliding door, a drawer, differently colored blocks, a button that toggles an LED, and a switch controlling a lightbulb. 
As shown in Fig.~\ref{fig:calvin_benchmark_illustration}, the environments differ in the textures of the desk, and the positions of all static elements including the sliding door, the drawer, the LED button, and the lightbulb switch. We conduct experiments under the most challenging ABC$\longrightarrow$D setting, i.e., training on data from environments  A, B, and C while testing in D.

\paragraph{Real-world Robot Experiments.}  We design three tasks for real-world evaluation: \texttt{Pick-Place Banana} (picking up a banana from the tabletop and placing it in a pan), \texttt{Close Laptop} (pushing the laptop's lid until it is completely closed), and \texttt{Disassembly} (removing the stick that is assembled in the slot). All tasks are executed with a FANUC LR Mate 200iD robot arm, as shown in Fig.~\ref{fig:real_robot_illustration}.
Each task is tested for 10 times, with the initial positions of objects randomized. For generalization evaluation, we consider two scenarios: (i) Novel Object: we change the color, texture, and shape of the manipulated object; 
(ii) Visual Distractor: we add irrelevant objects as distractors.

\subsection{Datasets} 

\paragraph{Training Latent Motion Tokenizer \& Pre-training Moto-GPT.} On the SIMPLER benchmark, we use a subset of Open-X-Embodiment (OXE)~\cite{vuong2023open} to train the Latent Motion Tokenizer and pre-train Moto-GPT, including 109k real-world trajectory videos from RT-1 Robot Action~\cite{brohan2022rt}, Bridge~\cite{walke2023bridgedata}, Task Agnostic Robot Play~\cite{rosete2022tacorl,mees23hulc2}, Jaco Play~\cite{dass2023jacoplay}, Cable Routing~\cite{luo2023multistage}, RoboTurk~\cite{mandlekar2019scaling}, NYU VINN~\cite{pari2021surprising}, Austin VIOLA~\cite{zhu2022viola}, Berkeley Autolab UR5~\cite{BerkeleyUR5Website}, and TOTO~\cite{zhou2023train} datasets across various embodiments. On CALVIN,  we use all the play videos from environments A, B, and C to train the Latent Motion Tokenizer. 35\% data (including 18k trajectory videos) with language annotations is used for autoregressive pre-training. 
The real-world experiments also use OXE data for pre-training.

\paragraph{Fine-tuning Moto-GPT.} On the SIMPLER benchmark, we use the 73k action-labeled real-world expert trajectories from  RT-1 Robot Action~\cite{brohan2022rt} to fine-tune the policy model. On the CALVIN benchmark, we use the 18k demonstration trajectories with language annotations and action labels from environments A, B, and C for fine-tuning. For real-world experiments, we collect 90 demonstration trajectories (30 for each task) with teleoperation for fine-tuning.
\\

\noindent Note that all the pre-training and fine-tuning data for SIMPLER come from the real world instead of simulation environments. This setting aims to study the model's transfer ability between real and simulation scenarios. On the other hand, the ABC-only setting for CALVIN training data aims to evaluate the model's zero-shot generalization capability to the unseen environment D.

\subsection{Compared Models}
\paragraph{SIMPLER.}
On the SIMPLER benchmark, we mainly compare Moto-GPT with five representative models pre-trained with Open-X-Embodiment datasets:
\begin{itemize}
    \item \textbf{RT-1-X}~\cite{brohan2022rt} uses a transformer backbone to output tokenized actions with a FiLM EfficientNet to fuse language and 6 history images into token inputs.
    
    \item \textbf{RT-2-X}~\cite{zitkovich2023rt} adapts the pre-trained large vision-language model (VLM), PaLI-X (55B), into a robot policy by casting tokenized actions into text tokens.
    \item \textbf{Octo-Base}~\cite{mees2024octo} employ a transformer architecture to process language and image tokens, with a diffusion-based action head to produce actions.
    \item \textbf{OpenVLA}~\cite{kim2024openvla} builds on a pre-trained Prismatic-7B VLM backbone for robot action prediction.
    \item \textbf{OpenVLA (fine-tuned)} further fine-tunes OpenVLA on the RT-1 Robot Action dataset~\cite{brohan2022rt}, despite its action-labeled pre-training data already contains this dataset.
\end{itemize}

\paragraph{CALVIN.}\label{sec:baseline}
On the CALVIN benchmark, we select the following baseline models that leverage pre-training strategies to improve robot manipulation performance:
\begin{itemize}
    \item \textbf{SuSIE}~\cite{black2024zeroshot} pre-trains an image editing model to generate the goal image, which is fed into a low-level policy for action prediction.
    \item \textbf{RoboFlamingo}~\cite{li2024visionlanguage} is a robot policy model adapted from OpenFlamingo, a large VLM pre-trained on extensive vision-language corpus.
    \item \textbf{GR-1}~\cite{wu2024unleashing} pre-trains a GPT-style transformer to directly predict the pixel values of a single-step future observation for each input observation.
    \item \textbf{MT-R3M}~\cite{wu2024unleashing} is a variation of GR-1, which leverages the pre-trained robot visual encoder R3M~\cite{nair2023r3m} to encode observation images.
\end{itemize}

\paragraph{Ablations of Moto-GPT.} We study the following variations of Moto-GPT as optional baselines:
\begin{itemize}
    \item \textbf{Moto w/o Motion Token} shares the same backbone with Moto-GPT but is trained from scratch on action-labeled robot data without latent motion tokens.
    \item \textbf{Moto-IML} undergoes the same pre-training stage as Moto-GPT. It keeps latent motion tokens in the input sequence but ignores the next-motion-token-prediction loss during the fine-tuning stage.  
    \item \textbf{Moto-DM} is pre-trained in the same way as Moto-GPT but completely discards latent motion tokens in the input sequence during fine-tuning.
\end{itemize}

\section{Training Details}
\subsection{Latent Motion Tokenizer}

The implementation details for the trainable modules of the Latent Motion Tokenizer are summarized in Table~\ref{tab:latent_motion_tokenizer_hyperparameters}. We use the hyperparameters listed in Table~\ref{tab:training_hyperparameters_latent_motion_tokenizer} to train this model on four 40G GPUs. 
To facilitate the learning of latent motion tokens, we downsample the original videos in the training dataset, ensuring that the visual motion between frames is sufficiently distinct. Specifically, for videos from the OXE data, we sample one frame every three frames (i.e., $\Delta t=3$) for videos from the RT-1 Robot-Action subset. The sampling rates for videos from other OXE subsets are adjusted based on the ratio of their fps to that of RT-1 Robot-Action videos. For human videos, $\Delta t$ is empirically set to 6. 
We train the Latent Motion Tokenizer for 350k steps. For videos from the CALVIN dataset, we adopt a sampling rate of one frame every five frames ($\Delta t=5$) and train the model for 150k steps. For real-world robot experiments, we fine-tune the Latent Motion Tokenizer pre-trained on OXE videos for another 500 steps on the newly collected real-world trajectory videos.

\begin{table}[t]
\caption{Implementation details of the Latent Motion Tokenizer.}
\small
    \centering
    \begin{tabular}{l c c}
    \toprule
    Component & Parameter & Value \\
    \midrule
    \multirow{4}{*}{M-Former} & num\_queries & 8 \\
    & num\_layers & 4 \\
    & hidden\_size & 768 \\
    & num\_heads & 12 \\
    \midrule
    \multirow{4}{*}{ViT Decoder} & patch\_size & 16 \\
    & num\_layers & 12 \\
    & hidden\_size & 768 \\
    & num\_heads & 12 \\
    \midrule
    \multirow{2}{*}{VQ Codebook} & num\_codes & 128 \\
    & latent\_dim & 32 \\

    \bottomrule
    \end{tabular}\label{tab:latent_motion_tokenizer_hyperparameters}
\end{table}

\begin{table}[!t]
\caption{Training hyperparameters for Latent Motion Tokenizer.}
\small
    \centering
    \begin{tabular}{l c}
    \toprule
    Parameter & Value  \\
    \midrule
    batch\_size & {256}  \\
    optimizer & {AdamW}  \\
    lr\_max & {1e-4}  \\
    lr\_schedule & {cosine decay}  \\
    weight\_decay & {1e-4}  \\
    warmup\_steps & {1000} \\
    \bottomrule
    \end{tabular}\label{tab:training_hyperparameters_latent_motion_tokenizer}
\end{table}

\begin{table}[!t]
\caption{Implementation details of Moto-GPT.}
\small
    \centering
    \begin{tabular}{l c c}
    \toprule
    Component & Parameter & Value \\
    \midrule
    \multirow{3}{*}{GPT backbone} & num\_layers & 12 \\
    & hidden\_size & 768 \\
    & num\_heads & 12 \\
    \midrule
    \multirow{2}{*}{Action Head} & num\_layers & 2 \\
    & hidden\_size & 384 \\
    \bottomrule
    \end{tabular}\label{tab:moto_gpt_hyperparameters}
\end{table}

\begin{table}[!t]
\caption{Training hyperparameters for Moto-GPT.}
\small
    \centering
    \begin{tabular}{l c}
    \toprule
    Parameter & Value \\
    \midrule
    batch\_size & {512} \\
    optimizer & {AdamW} \\
    lr\_max & {1e-4} \\
    lr\_schedule & {cosine decay} \\
    weight\_decay & {1e-4}\\
    warmup\_epochs & {1} \\
    \bottomrule
    \end{tabular}\label{tab:training_hyperparameters_moto_gpt}
\end{table}

\subsection{Moto-GPT}
We present the implementation details of Moto-GPT in Table~\ref{tab:moto_gpt_hyperparameters}, where the Action Head is included only during the fine-tuning phase. Moto-GPT handles a maximum video length of three frames, and the video downsampling rate applied during both the pre-training and fine-tuning stages is consistent with the rate used for training the Latent Motion Tokenizer.
When fine-tuning Moto-GPT across different benchmarks, the number of action query tokens inserted after the latent motion tokens at each time step varies. Specifically, for the SIMPLER benchmark, we insert three action query tokens, whereas for the CALVIN benchmark, we insert five.
For pre-training, Moto-GPT is trained for 10 epochs using eight 40G GPUs, with the relevant hyperparameters outlined in Table~\ref{tab:training_hyperparameters_moto_gpt}. The hyperparameters for fine-tuning remain consistent with those used during pre-training, with the exception of the number of epochs. During fine-tuning, Moto-GPT is trained for three epochs on the RT1-Robot-Action dataset and 18 epochs on the CALVIN dataset, utilizing four 40G GPUs. For real-world experiments, we start with the same pre-trained checkpoint of Moto-GPT as adopted for the SIMPLER benchmark. We further pre-train it with a combination of OXE videos and the 90 newly collected trajectory videos for 5 epochs. Then we fine-tune it with real robot action labels for 20 epochs.

\section{Details of Experiments}

\begin{figure}[!htbp]
    \centering
        \vspace{-5pt}
    \includegraphics[width=0.5\textwidth]{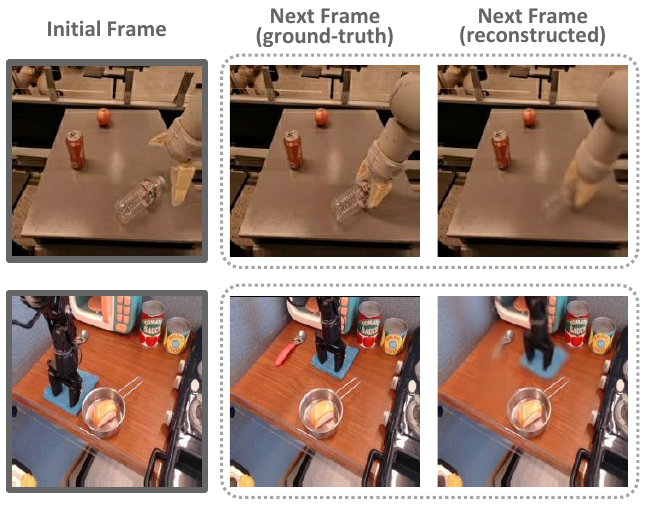}
    \caption{
    Qualitative examples of reconstruction results, where discrete motion tokens obtained from the Latent Motion Tokenizer based on the initial and next frame, are fed into the decoder along with the initial frame to reconstruct the target frame.
    }
    \label{fig:recons_quality}
        \vspace{-5pt}
\end{figure}

\begin{figure*}[!htbp]
    \centering
    \includegraphics[width=1.0\textwidth]{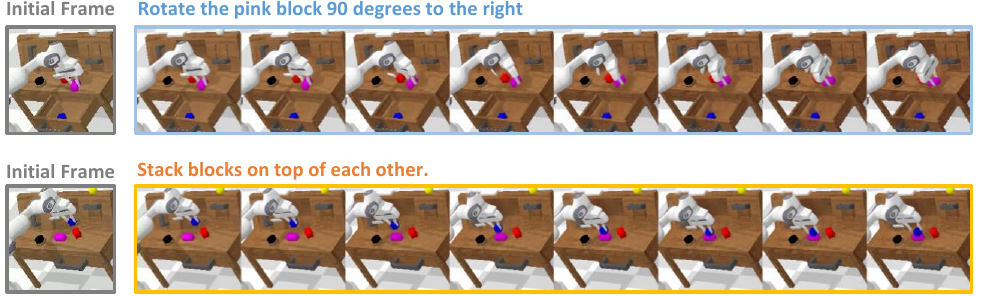}
    \caption{Predicted video trajectories by the pre-trained Moto-GPT for CALVIN tasks reflecting delicate robot actions.}
    \label{fig:prediction_for_complex_actions}
\end{figure*}

\begin{table}[!htbp]
\caption{
Top-K motion token prediction accuracy of Moto-GPT in predicting ground-truth latent motion tokens from a 128-size codebook on the validation splits of the pre-training datasets. 
}
\small
    \centering
    \begin{tabular}{l c c c}
    \toprule
    Dataset & Top-5 & Top-10 & Top-20 \\
    \midrule
    Oepn-X-Embodiment & 0.521 & 0.698 & 0.853 \\
    Calvin (ABC$\longrightarrow$D) & 0.298 & 0.518 & 0.768 \\
    \bottomrule
    \end{tabular}\label{tab:topk_token_acc}
\end{table}

\section{Limitations \& Future Work}

This paper introduces Moto, a novel method that uses latent motion tokens as a ``language'' interface to bridge generative pre-training on video data with precise robot control. Moto opens several exciting avenues for future work. 

Firstly, Moto demonstrates the feasibility of learning a unified language to interpret diverse visual dynamics from videos, eliminating the need for hardware-specific action labels. The latent motion trajectories tokenized from videos provide a rich resource for models to learn motion priors closely related to low-level actions. 
While we currently mainly use robot videos to train the Latent Motion Tokenizer, the learned latent motion tokens demonstrate the potential to produce consistent visual motions across varied contexts and embodiments. 
Notably, our preliminary experiments with SSV2 videos show promising results in human-to-robot motion transfer via latent motion tokens. We believe a similar approach could be applied to model more complex human motions, enabling robots to learn a wealth of world knowledge from Internet-scale videos.

Besides, the Moto-GPT pre-trained on videos tokenized into latent motion token sequences and fine-tuned on action-labeled trajectories, effectively transfers motion priors learned from (even human) videos to actual robot action prediction. This is particularly beneficial in low-resource scenarios. 
Future work could scale up pre-training video data and optimize fine-tuning to improve model performance on downstream robot tasks further.

Moreover, while Moto is primarily utilized to enhance imitation learning for robot manipulation tasks, it shows potential as a reward model for measuring trajectory rationality and as a vivid environment simulator. Future research could explore Moto's use in improving the robustness of reinforcement learning agents and extending its application to a wider range of robotic tasks, such as navigation and locomotion, to develop a more versatile robot action policy.

\end{document}